\documentclass[sigconf]{acmart}

\setcopyright{none}
\renewcommand\footnotetextcopyrightpermission[1]{}

\usepackage{multirow}
\usepackage[table,xcdraw]{xcolor}

\usepackage{subcaption}

\begin{document}

\title{\includegraphics[height=1em]{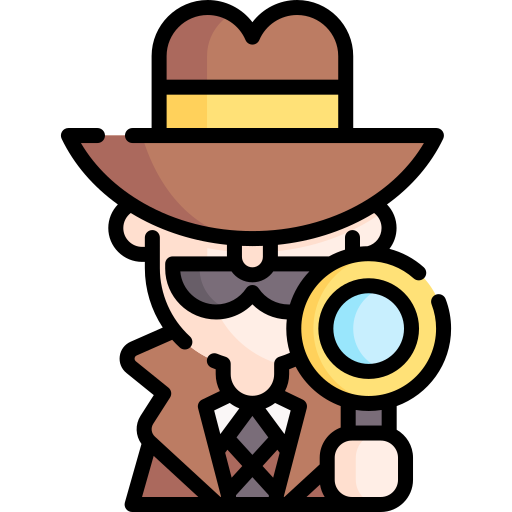} VisRAG2.0: Mitigating Visual Hallucinations via Evidence-Guided Multi-Image Reasoning in Visual Retrieval-Augmented Generation}




\author{Yubo Sun}
\authornote{Equal contribution.}
\affiliation{
  \institution{Peking University}
  \country{China}
}
\email{boggysyb@gmail.com}

\author{Chunyi Peng}
\authornotemark[1]
\affiliation{
  \institution{Northeastern University}
\country{China}
}
\email{hm.cypeng@gmail.com}

\author{Yukun Yan}
\authornote{Corresponding author.}
\affiliation{
  \institution{Tsinghua University}
  \country{China}
}
\email{yanyk.thu@gmail.com}

\author{Shi Yu}
\affiliation{
  \institution{Tsinghua University}
  \country{China}
}
\email{yushi17@foxmail.com}

\author{Zhenghao Liu}
\affiliation{
  \institution{Northeastern University}
  \country{China}
}
\email{liuzhenghao@mail.neu.edu.cn}

\author{Sen Mei}
\affiliation{
  \institution{Tsinghua University}
  \country{China}
}
\email{meisen2025@gmail.com}

\author{Chi Chen}
\affiliation{
  \institution{Tsinghua University}
  \country{China}
}
\email{chenchithu@gmail.com}

\author{Maosong Sun}
\authornotemark[2]
\affiliation{
  \institution{Tsinghua University}
  \country{China}
}
\email{sms@tsinghua.edu.cn}

\begin{abstract}

Visual Retrieval-Augmented Generation (VRAG) has emerged as a promising paradigm for equipping Vision-Language Models (VLMs) with external visual evidence, enabling them to go beyond parametric knowledge when answering visually grounded questions. However, in such multi-image settings, VLMs still often suffer from visual hallucinations and struggle to accurately identify the question-relevant evidence needed for reliable reasoning. Existing methods usually lack an explicit cross-image evidence collection process, and also provide limited credit assignment when jointly optimizing perception and reasoning.
To address this issue, we propose EVisRAG, an evidence-guided visual retrieval-augmented framework for multi-image reasoning. EVisRAG first observes the retrieved images, records question-relevant visual evidence from each image, and then performs reasoning and answer generation based on the aggregated evidence. We further introduce RS-GRPO, which aligns reward signals with token spans from different stages, improving training stability and strengthening the joint optimization of evidence localization and reasoning. 
Experiments on multiple visual question answering benchmarks show that EVisRAG consistently outperforms the backbone VLM by an average of about 19\%, while substantially reducing visual hallucinations. These results demonstrate that explicit evidence collection and scoped reward design are effective for improving visual grounding and reasoning reliability in multi-image settings. Codes and data are available at \url{https://github.com/OpenBMB/VisRAG}
\end{abstract}

\maketitle
\section{Introduction}

Retrieval-Augmented Generation (RAG) equips Large Language Models (LLMs) with external knowledge retrieval to provide task-relevant context and mitigate hallucinations caused by limited parametric knowledge~\citep{lewis2020retrieval, liu2026knowledge}. 
However, much real-world knowledge is inherently non-textual, residing in modalities such as images, tables, and complex document layouts. 
Text-centric preprocessing pipelines that rely on image captioning or OCR to linearize these signals discard essential visual and spatial structure, limiting the model’s ability to access and reason over information originally present in images or document pages~\citep{zhang2024map}.

To address this limitation, Visual RAG (VRAG)~\citep{yu2024visrag,faysse2407colpali} retrieves document page snapshots as units, preserving visual and spatial cues so VLMs can read evidence directly from images. 
Recent variants couple retrieval with reinforcement learning, inserting retrieved images into intermediate reasoning steps so the model can derive the correct answer from pixels rather than text alone~\citep{peng2026mixture, wu2025mmsearch}. 
Despite these gains, many methods still transplant text-based RAG practices into the visual domain and neglect modality-specific capabilities such as perceiving information relevant to the question from images. 
Such a deficiency in visual perception often exacerbates visual hallucinations.
In particular, models may fail to correctly attend to and utilize relevant visual evidence present in the images, leading to incorrect answers, or may erroneously perceive and reason about visual content that does not actually exist, resulting in spurious or hallucinated conclusions. Some works introduce perception-oriented actions or auxiliary agents to mitigate visual hallucinations~\citep{wang2025vrag,wang2025vidorag}, which improves attention to visual detail but increases architectural complexity and computational cost, complicating end-to-end training and later reconfiguration.

\begin{figure*}[t!]
  \centering
  \includegraphics[width=0.95\textwidth]{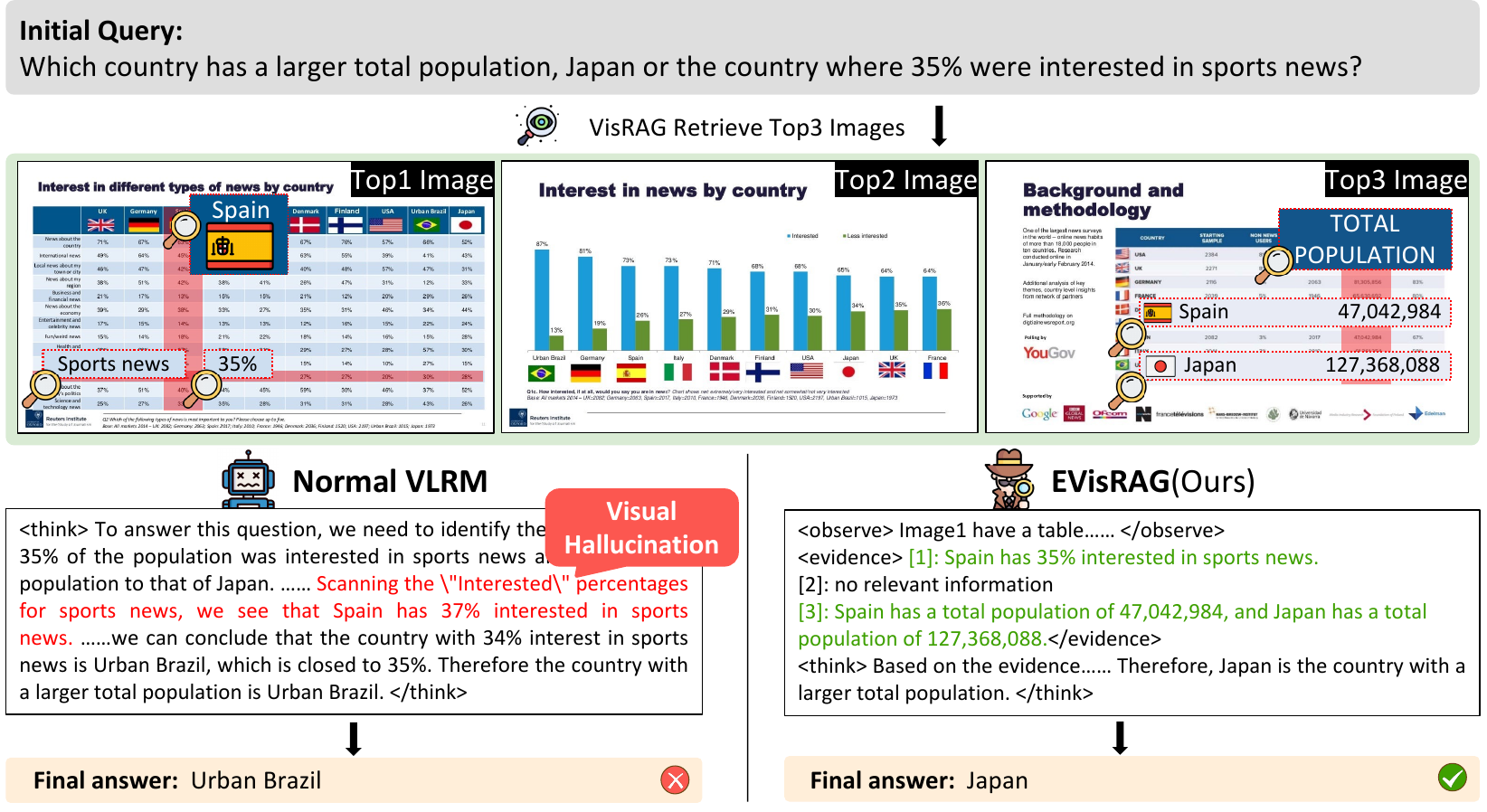}  
  \caption{Comparison of normal vision-language reasoning model (VLRM) and EVisRAG}
  \label{fig:intro}
\end{figure*}
To bypass these complexities, recent advances in vision-language reasoning models (VLRMs) have introduced promising strategies for enhancing visual perception on a single image during the reasoning process~\citep{shen2025vlm,xu2025mixed} by incorporating auxiliary rewards related to visual perception. 
Although these VLRMs perform well on single-image inputs, VRAG often retrieves multiple images, requiring cross-image localization and integration of visual evidence. 
Current methods lack a built-in per-image evidence collection and instead rely on external tools or agents, which increases complexity and instability. 
In addition, current VLRM training strategies typically optimize perception and reasoning with mixed rewards, overlooking the effective scope and objective differences of each signal, which blurs credit assignment and causes interference.

Motivated by these challenges, we propose \textbf{E}vidence-guided \textbf{Vis}ion \textbf{R}etrieval-\textbf{a}ugmented \textbf{G}eneration (EVisRAG) to equip VLMs with precise visual perception in multi-image scenarios.
As illustrated in Figure~\ref{fig:intro}, EVisRAG conducts a linguistic observation phase that sequentially gathers evidence from retrieved images, maintaining focus on them, and then performs reasoning on the collected evidence to derive the correct answer. 
To train EVisRAG effectively, we introduce Reward-Scoped Group Relative Policy Optimization (RS-GRPO), which applies fine-grained rewards directly to in-scope tokens corresponding to visual evidence collection or reasoning.
Unlike conventional joint optimization with mixed rewards, where perception and reasoning signals interfere and lead to unstable or suboptimal training, RS-GRPO explicitly scopes rewards to their effective regions, enabling precise credit assignment and stable optimization.
Without such reward scoping, jointly optimizing visual perception and reasoning in multi-image settings becomes unstable, as gradients from heterogeneous objectives compete across irrelevant tokens.
Experiments on different VQA tasks demonstrate the effectiveness of EVisRAG, showing substantial improvements over different VRAGs. 
Powered by RS-GRPO, EVisRAG can precisely identify question-relevant evidence image by image and reason over the recorded cues to produce grounded answers, much like a detective assembling evidence. 
Moreover, EVisRAG demonstrates stronger visual perception and higher answer accuracy among other baselines, confirming that richer visual perception improves the ability of question understanding and alleviates visual hallucinations.

\section{Related Work}

Early research on retrieval-augmented generation (RAG) equips large language models (LLMs) with retrievers over curated corpora to provide task-relevant context and mitigate hallucinations~\citep{lewis2020retrieval, asai2024self}. However, a substantial portion of real-world knowledge is non-textual, residing in images, tables, and documents with complex layouts. Pipelines that first linearize such signals via captioning or optical character recognition and then provide only text to the model often discard critical visual and spatial cues, which degrades performance on downstream reasoning tasks~\citep{zhang2024map}. To address this limitation, VisRAG~\citep{yu2024visrag} and Colpali~\citep{faysse2407colpali} introduce Visual Retrieval-Augmented Generation (VRAG), which treats document page snapshots as retrieval units and enables vision–language models to directly read evidence from images.

Nevertheless, regardless of the modality of the retrieved context, existing RAG and VRAG models remain prone to hallucinations and reasoning failures. In particular, models may confidently produce incorrect answers despite the presence of relevant evidence, become distracted by irrelevant or weakly related information, or fail to accurately extract and reason over key facts from long contextual inputs~\citep{mishra2024fine, cuconasu2024power, hsieh2024found}. Building on this observation, retrieval-augmented reasoning approaches~\citep{shao2024deepseekmath, rafailov2023direct, schulman2017proximal, li2025search, song2025r1} acquire and exploit evidence at intermediate reasoning steps, using it to guide subsequent inference and reduce hallucinations arising from misused textual context. Despite these advances, many methods directly transplant text-centric RAG practices to the visual domain and insufficiently account for modality-specific challenges such as cross-image grounding, layout-aware reading, and region-level attention, resulting in unstable perception across multiple images~\citep{wang2025vidorag,wang2025vrag}.

Recent advances in vision and language reasoning models (VLRMs) have introduced effective strategies for strengthening visual perception during reasoning. Vision-R1~\citep{zhan2025vision}, MM-Eureka~\citep{meng2025mm}, Ocean-R1~\citep{ming2025oceanr1}, ThinkLite-VL~\citep{wang2025sota}, and OpenVLThinker~\citep{deng2025openvlthinker} show that directly applying GRPO, sometimes even without supervised fine-tuning, substantially promotes the emergence of chain of thought reasoning and can elicit ``aha" moments. VLM-R1~\citep{shen2025vlm} and Mixed-R1~\citep{xu2025mixed} further improve perceptual grounding by augmenting answer correctness signals with auxiliary perception rewards, encouraging better use of image information. However, in VRAG settings that require reasoning over semantically rich content from multiple images, the remaining limitations in perceptual grounding often lead to misinterpretation of visual evidence, which in turn undermines the validity of the overall reasoning process.

\section{Methodology}

\begin{figure*}[t!]
  \centering
  \includegraphics[width=0.95\textwidth]{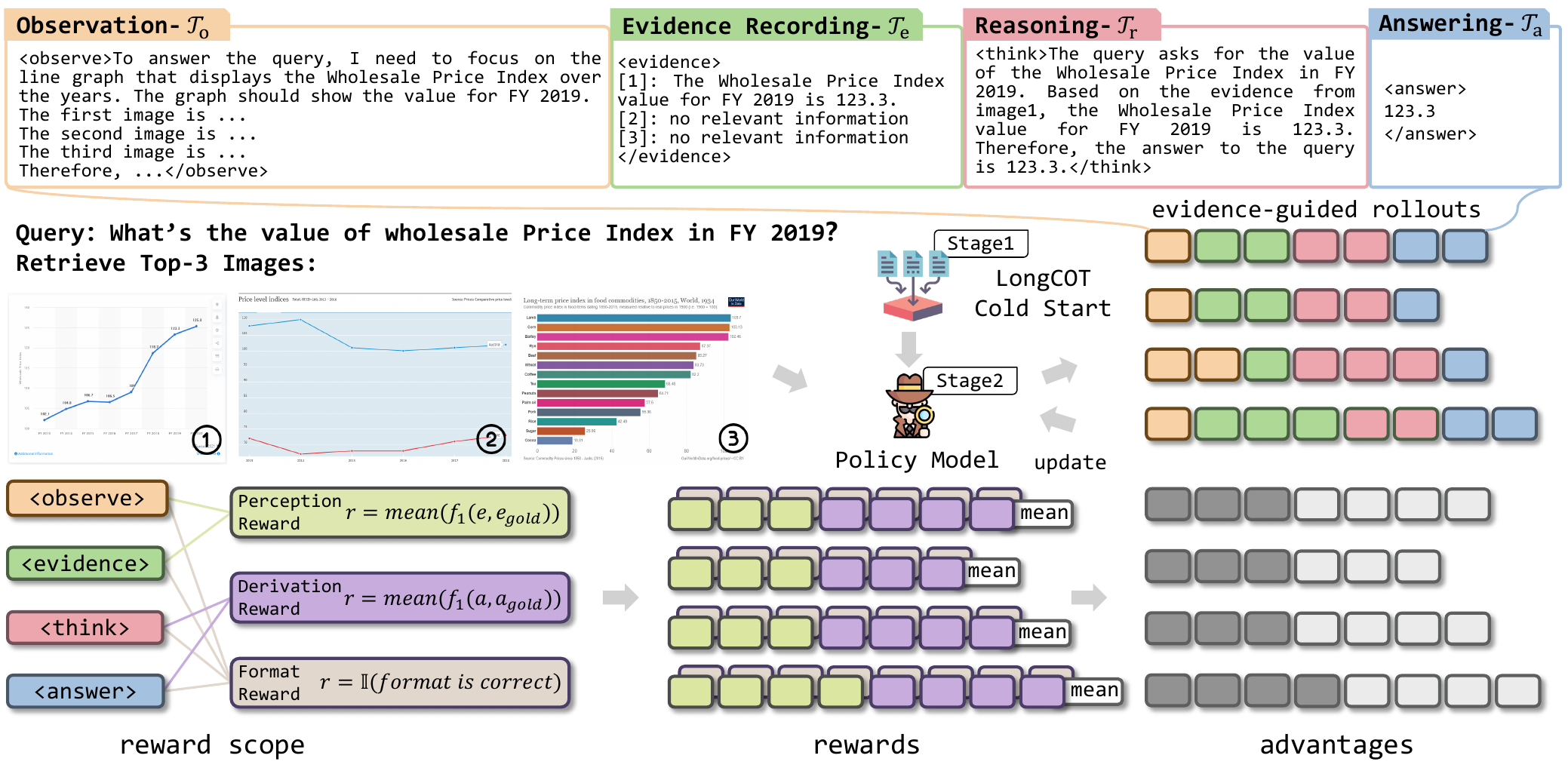}  
  \caption{Overall framework of EVisRAG. Followed by the query and top-3 retrieved document pages, EVisRAG outputs four token scopes: observation, evidence recording, reasoning, and answering. RS-GRPO assigns three fine-grained rewards to scope-specific tokens. In-scope rewards are then averaged and group-normalized to obtain token advantages for policy updates.
}
  \label{fig:method}
\end{figure*}

This section presents EVisRAG, a framework that enables vision language models (VLMs) to reason over multiple images with rich visual evidence. We first outline its evidence-guided reasoning process, including observation, evidence recording, and answer reasoning (Section~\ref{sec:overview}). We then introduce Reward-Scoped Group Relative Policy Optimization (RS-GRPO), which enhances fine-grained perceptual grounding during reasoning (Section~\ref{sec:GRTPO}).

\subsection{Overview of EVisRAG}
\label{sec:overview}

Given a query $q$ and corpus $\mathcal{D}$ of document-page snapshots, EVisRAG performs step-by-step evidence-guided reasoning to retrieve and localize visual evidence and produce the final answer $a$:
\begin{equation}
    (q, \mathcal{D})\xrightarrow{\text{EVisRAG}}a,
\end{equation}
where $q$ is an open-domain question and corpus $D$ indexes page-level images, providing visual evidence relevant to $q$ for a VLM to exploit.

\textbf{Retrieval.} 
The first stage of EVisRAG aims to retrieve a set of pages from a large document page corpus $\mathcal{D}$, given a query $q$. Following the VisRAG~\citep{yu2024visrag} retrieval paradigm, we obtain
\begin{equation}
\mathcal{D}_\mathcal{R} = \text{VisRAG-Ret}(q, \mathcal{D}, k),
\end{equation}
where the candidate set $\mathcal{D}_\mathcal{R}\subset\mathcal{D}$ contains the top-$k$ document pages relevant to the question $q$.

\textbf{Visual Perception.}
After gathering candidates $\mathcal{D}_{\!R}=\{d_i\}_{i=1}^{k}$ from $\mathcal{D}$, EVisRAG sequentially observes these pages and produces a coarse, page-aware observation $y_{o}$:
\begin{equation}
    P\!\left(y_{o}\mid q,\mathcal{D}_{\!R}\right)
    = \prod_{t=1}^{|\mathcal{T}_o|} P\!\left(y_{o,t}\mid y_{o,<t},\, q,\, \mathcal{D}_{\!R}\right),
\end{equation}
where $\mathcal{T}_o$ denotes the output token sequence of the observation stage, $|\mathcal{T}_o|$ represents the length of this sequence, and $y_{o,t}$ denotes to the $t$-th generated observation token. The objective of this stage is to establish an accurate understanding of the visual content across multiple images, thereby providing a foundation for subsequent evidence recording.

Conditioned on $q$ and $y_{o}$, EVisRAG then records evidence from each page by generating per-image evidence sequences $y_{e}^{(i)}$:
\begin{equation}
\begin{aligned}
P(y_{e} \mid q, y_{o}, \mathcal{D}_R)
&= \prod_{i=1}^{k} \prod_{t=1}^{|\mathcal{T}_e^{(i)}|}
P\bigl(y_{e,t}^{(i)} \mid y_{e,<t}^{(i)},
q, y_{o}, d_i \bigr).
\end{aligned}
\end{equation}
where $y_{e}=\{y_{e}^{(i)}\}_{i=1}^{k}$, $\mathcal{T}_e$ denotes the output token sequence of the evidence recording stage, $|\mathcal{T}_e^{(i)}|$ is the length of the evidence sequence for the retrieved document page $d_i$, and $y_{e,t}^{(i)}$ is its $t$-th token.

\textbf{Answer Reasoning.}
After visual perception, EVisRAG conducts detective-style reasoning over the perceived information
$y_{p}=\{y_{o},\, y_{e}\}$: it distills leads from the recorded evidence, formulates and tests hypotheses across pages, cross-checks contradictions, and organizes a coherent reasoning trajectory $y_{r}$:
\begin{equation}
\begin{aligned}
P(y_{r} \mid q, y_{p}, \mathcal{D}_R)
&= \prod_{t=1}^{|\mathcal{T}_r|}
P\bigl(y_{r,t} \mid  y_{r,<t},\, q,\,
y_{p},\, \mathcal{D}_R\bigr).
\end{aligned}
\end{equation}
where $\mathcal{T}_r$ denotes the output token sequence of the reasoning stage, $|\mathcal{T}_r|$ is the length of the reasoning sequence and $y_{r,t}$ denotes its $t$-th token.

Conditioned on $q$, $y_{p}$, and $y_{r}$, EVisRAG then produces the final answer sequence $y_{a}$:

\begin{equation}
\begin{aligned}
P(y_{a} \mid q, y_{p}, y_{r}, \mathcal{D}_R)
= \prod_{t=1}^{|\mathcal{T}_a|}
P\Bigl(
y_{a,t}
\,\Bigm|\, y_{a,<t}, q,\, y_{p},\, y_{r},\, \mathcal{D}_R
\Bigr).
\end{aligned}
\end{equation}
where $\mathcal{T}_a$ denotes the output token sequence of the answering stage, $|\mathcal{T}_a|$ is the answer length and $y_{a,t}$ is its $t$-th token.

\subsection{Optimizing VLMs for Evidence-guided Reason via RS-GRPO}
\label{sec:GRTPO}
To enhance EVisRAG’s ability to accurately record evidence from multiple images and reason based on that evidence, we employ a two-stage training as shown in Figure~\ref{fig:method}. In the first stage, we apply supervised fine-tuning (SFT) as a cold start. In the second stage, we introduce Reward-Scoped Group Relative Policy Optimization (RS-GRPO), which extends GRPO~\citep{shao2024deepseekmath} to jointly optimize perception and reasoning ability of VLMs, with fine-grained rewards applied to their corresponding reward scopes.

\textbf{Reward Scopes.}

To evaluate model outputs while encouraging the evidence-guided reasoning paradigm, we adopt three fine-grained rewards in a coordinated scheme. The format reward $R_{\text{format}}$ enforces adherence to an evidence-guided reasoning paradigm by requiring the model to observe, record evidence, reason, and answer in a disciplined order, making intermediate steps explicit and supervision stable. The perception reward $R_{\text{perception}}$ checks whether question-relevant regions are correctly localized and summarized for each image based on the ground truth evidence generated by a larger VLM, and allows for explicit no relevant information when evidence is absent. The derivation reward $R_{\text{derivation}}$ evaluates whether the model derives the correct final answer from its visual perception, ensuring the reasoning is grounded in the observed and recorded evidence. More details of the reward design are shown in the Appendix.


To jointly train the perception and reasoning ability of VLMs, we introduce Reward Scopes, which route supervision to scope-specific tokens to sharpen credit assignment, reduce interference, and stabilize training. Let $\mathcal{M}(t)$ denote the set of reward channels applicable to the token at position $t$. The output sequence is segmented by special tokens into four scopes: the observation scope $\mathcal{T}_o$, the evidence recording scope $\mathcal{T}_e$, the reasoning scope $\mathcal{T}_r$, and the answering scope $\mathcal{T}_a$. Rewards are effective only when they are meaningful, as $y_{p}$ supervises tokens in the visual perception scope: $\mathcal{T}_o$ and $\mathcal{T}_e$, guiding the model to summarize the right visual regions. $R_{\text{derivation}}$ supervises tokens in the derivation scope: $\mathcal{T}_r$ and $\mathcal{T}_a$, encouraging the model to derive the correct final answer from what was perceived. $R_{\text{format}}$ applies to all tokens and keeps the evidence-guided workflow explicit and stable. Formally, we define the reward–scope mapping as:
\begin{equation}
\mathcal{M}(t)=
\begin{cases}
\{\,R_{\text{perception}},\,R_{\text{format}}\,\} & t\in \mathcal{T}_o \cup \mathcal{T}_e\\
\{\,R_{\text{derivation}},\,R_{\text{format}}\,\} & t\in \mathcal{T}_r \cup \mathcal{T}_a
\end{cases}.
\end{equation}

For the $i$-th sampled output and its token at position $t$, let $R^{(m),i}_{t}$ denote the
score from reward channel $m\in\mathcal{M}(t)$.Each reward channel is broadcast to all tokens within its effective scope before in-scope averaging and group normalization, and the scope-aggregated token reward is the
mean over its in-scope channels:
\begin{equation}
\bar{R}^{\,i}_{t}=\frac{1}{|\mathcal{M}(t)|}\sum_{m\in\mathcal{M}(t)} R^{(m),i}_{t}.
\end{equation}

\textbf{RS-GRPO objective.}
To train both visual perception and reasoning, EVisRAG adopts an RS-GRPO objective that
explicitly computes token advantages under reward scopes. Given a group of $G$ sampled
outputs, the token-level advantage is
\begin{equation}
    \hat A^{\,i}_{t} \;=\;
\frac{\bar{R}^{\,i}_{t} \;-\;\mathrm{mean}\bigl(\{\bar{R}^{\,1}_{t}, \bar{R}^{\,2}_{t}, \ldots, \bar{R}^{\,G}_{t}\}\bigr)}
     {\mathrm{std}\bigl(\{\bar{R}^{\,1}_{t}, \bar{R}^{\,2}_{t}, \ldots, \bar{R}^{\,G}_{t}\}\bigr)},
\end{equation}
where $i$ indexes the $i$-th sample in the group, and $G$ is the group size. We incorporate the resulting token-level advantages into DAPO~\cite{yu2025dapo} to enhance exploration diversity and training stability, and optimize the model by minimizing the following objective:

\begin{equation}
\begin{aligned}
\mathcal{L}_{\text{RS-GRPO}}(\theta)
&= - \frac{1}{\sum_{i=1}^{G} |o^i|}
\sum_{i=1}^{G} \sum_{t=1}^{|o^i|}
\min \Big(
r_t^i(\theta)\hat{A}_t^i, \\
&\quad \text{clip}\big(
r_t^i(\theta),
1-\epsilon_{\text{low}},
1+\epsilon_{\text{high}}
\big)\hat{A}_t^i
\Big),
\end{aligned}
\end{equation}
where $o^{i}$ is the $i$-th sampled output sequence,
$r^{\,i}_{t}(\theta)$
is the importance ratio, and $\epsilon_{\mathrm{low}},\epsilon_{\mathrm{high}}$ are the
lower and upper clipping thresholds.

\section{Experimental Methodology}
This section describes the datasets, baselines, evaluation metrics, and implementation details.


\textbf{Evaluation Datasets.} We evaluate EVisRAG on five visual question answering (VQA) benchmarks covering diverse document types. ChartQA~\citep{masry2022chartqa} and InfoVQA~\citep{mathew2022infographicvqa} for figure understanding, DocVQA~\citep{tito2023hierarchical} for industrial documents, SlideVQA~\citep{tanaka2023slidevqa} for presentation slides, and ViDoSeek~\citep{wang2025vidorag} for multi-document scenarios.
For each query, we utilize VisRAG-Ret~\citep{yu2024visrag} to retrieve the top-3 relevant images as context. Subsequently, each question is categorized according to whether the retrieved context provides sufficient information to answer the question with sufficient context or with insufficient context. More details of the test datasets are provided in the Appendix.

\textbf{Training Data Construction.} We collect 30,000 QA samples from the training sets of ChartQA and InfoVQA, and partition them into SFT and GRPO subsets with an 8:2 ratio. For each query, we employ a VisRAG retriever to retrieve the top-1 to top-5 images as visual context for data augmentation. To construct reliable reasoning trajectories, we employ Qwen2.5-VL-72B~\citep{bai2025qwen2} to generate candidate chains of thought, and retain only those samples whose final answers are correct and whose evidence IDs precisely match the truly relevant images. This procedure effectively filters out noisy or hallucinated supervision and prevents the student model from inheriting systematic errors from the teacher. We further conduct a lightweight manual inspection on a subset of the collected data, including 75 answerable cases and 25 unanswerable cases. The inspection verifies that the teacher-generated evidence is consistently grounded in the corresponding image content and reveals no noticeable quality issues. Following~\citet{Polaris2025}, we further exclude samples that can already be solved by the SFT-trained model, thereby focusing subsequent training on more challenging examples. This process yields a final training set of 60,000 SFT samples and 4,000 GRPO samples. Detailed data construction procedures are provided in the Appendix.

\textbf{Baselines.} All baselines use VisRAG-Ret for retrieval. For each query, we fetch the top-$3$ documents, then the model answers using the retrieved images and the original question.

We compare three groups. General VLMs include Qwen2.5-VL-7B, Qwen2.5-VL-32B~\citep{bai2025qwen2}, MiMo-VL-7B-RL~\citep{coreteam2025mimovltechnicalreport} and Qwen3-VL-8B~\cite{bai2025qwen3}. VLRMs trained on Qwen2.5-VL-7B-Instruct include Vision-R1-7B~\citep{zhan2025vision}, Ocean-R1-7B~\citep{ming2025oceanr1}, ThinkLite-VL-7B~\citep{wang2025sota} and OpenVLThinker-7B~\citep{deng2025openvlthinker}. VRAG methods with the same backbone include ViDoRAG~\cite{wang2025vidorag}, R1-Router~\citep{peng2025learning}, MMSearch-R1~\citep{wu2025mmsearch}, and VRAG-RL~\citep{wang2025vrag}. More implementation details of the baseline methods are provided in the Appendix.

\textbf{Evaluation Metrics.} 
Due to inherent limitations in retrieval, the selected context may or may not provide sufficient information to answer the query. To rigorously assess both the perceptual and reasoning capabilities of the model while mitigating visual hallucination, we categorize each query-context pair into two types: sufficient context and insufficient context~\citep{joren2024sufficient}.

For query-context pairs with sufficient context, where the retrieved images provide enough evidence to support the answer, we adopt the original reference answer as the ground truth and evaluate model performance using both \textit{Accuracy} and \textit{F1 Score}.

For query-context pairs with insufficient context, where the retrieved images do not provide enough evidence to support a correct answer, the model is expected to abstain from answering. Instead of relying on strict string matching, we use GPT-5~\cite{singh2025openai} to evaluate whether the model appropriately refuses to answer due to insufficient contextual evidence.

Additional implementation details for the baseline methods are provided in the Appendix

\textbf{Implementation Details.} For fair comparison, we adopt Qwen2.5-VL-7B~\citep{bai2025qwen2} as the backbone of EVisRAG, as most baselines are also built on the same backbone. We use LLaMA-Factory~\citep{zheng2024llamafactory} and Easy-R1~\citep{zheng2025easyr1} for open-sourcing the training framework that we used for SFT and GRPO. All experiments were executed on GPU clusters with computational capabilities comparable to NVIDIA A100 80GB GPUs. Further details on the hyperparameters that we used for SFT and GRPO are provided in the Appendix.

\begin{table*}[t!]
    \centering
    \small 
    \setlength{\tabcolsep}{8pt}
    \renewcommand{\arraystretch}{1.0}
    \caption{Overall Performance of EVisRAG and Baselines. \textbf{Bold} denotes the highest value, and \underline{underline} denotes the second highest value. \textit{}{F1} Score is reported only on the sufficient-context subset.}
    \begin{tabular}{lcccccccccccc} \toprule 
   \multirow{3}{*}{\textbf{Methods}} & \multicolumn{4}{c}{\textbf{In Distribution}} & \multicolumn{6}{c}{\textbf{Out of Distribution}} & \multicolumn{2}{c}{\multirow{2}{*}{\textbf{Average}}} \\ \cmidrule(lr){2-5} \cmidrule(lr){6-11} 
      ~ & \multicolumn{2}{c}{\textbf{ChartQA}} & \multicolumn{2}{c}{\textbf{InfoVQA}} & \multicolumn{2}{c}{\textbf{DocVQA}} & \multicolumn{2}{c}{\textbf{SlideVQA}} & \multicolumn{2}{c}{\textbf{ViDoSeek
      }} & ~ & ~\\ \cmidrule(lr){2-5} \cmidrule(lr){6-11} \cmidrule(lr){12-13}
      ~ & Acc & F1 & Acc & F1 & Acc & F1 & Acc & F1 & Acc & F1 & Acc & F1 \\ \midrule
      \rowcolor{gray!20} \multicolumn{13}{l}{\textbf{General VLMs}} \\ \midrule

Qwen2.5-VL-7B & 56.32 & 14.99 & 56.41 & 35.31 & 53.81 & 25.59 & 62.23 & 30.16 & 42.38 & 29.55 & 54.23 & 27.12 \\ 
MiMo-VL-7B-RL & 66.32 & 20.45 & 71.87 & 22.75 & 81.22 & 30.41 & 83.63 & 14.31 & 51.84 & 24.28 & 70.98 & 22.44 \\
Qwen2.5-VL-32B & 70.48 & 52.42 & 78.41 & 52.43 & 84.26 & 61.69 & 79.86 & 50.44 & 48.16 & 42.67 & 72.23 & 51.93 \\
Qwen3-VL-8B & 67.92 & 51.49 & \underline{79.13} & \underline{78.50} & \underline{84.43} & \underline{84.79} & \underline{79.50} & 74.67 & \textbf{52.45} & \underline{65.78} & \underline{72.69} & 71.05 \\ \midrule
\rowcolor{gray!20} \multicolumn{13}{l}{\textbf{VLRMs}} \\ \midrule
Vision-R1 & 57.68 & 19.61 & 33.57 & 22.20 & 44.67 & 23.56 & 58.63 & 30.42 & 40.89 & 26.97 & 47.09 & 24.55 \\
ThinkLite-VL-7B & 60.64 & 30.93 & 62.12 & 52.48 & 72.25 & 53.43 & 73.74 & 51.20 & 48.25 & 41.94 & 63.40 & 46.00 \\
OpenVLThinker & 69.44 & 63.07 & 70.61 & 70.74 & 77.16 & 78.33 & 73.02 & 73.44 & 45.80 & 60.08 & 67.21 & 69.13 \\ 
Ocean-R1-7B & 69.00 & 51.92 & 70.33 & 69.56 & 80.54 & 78.14 & 75.90 & 73.02 & 50.70 & 61.22 & 69.29 & 66.77 \\ \midrule

\rowcolor{gray!20} \multicolumn{13}{l}{\textbf{VRAGs}} \\ \midrule
MMSearch-R1 & 62.00 & 33.31 & 52.92 & 47.34 & 59.73 & 47.20 & 64.21 & 47.93 & 44.05 & 44.00 & 56.58 & 43.96 \\
VRAG-RL & 56.08 & 10.84 & 57.94 & 10.16 & 56.35 & 16.28 & 73.56 & 15.03 & 43.26 & 18.09 & 57.44 & 14.08 \\
ViDoRAG & 61.76 & 51.31 & 53.48 & 51.38 & 76.31 & 78.18 & 65.47 & 64.30 & 45.88 & 58.73 & 60.58 & 60.78 \\
R1-Router & 56.40 & 1.67 & 68.94 & 1.74 & 72.08 & 3.71 & 75.18 & 2.68 & 46.67 & 4.50 & 63.85 & 2.86 \\ \midrule
\rowcolor{gray!20} \multicolumn{13}{l}{\textbf{EVisRAG(ours)}} \\ \midrule
\textbf{EVisRAG-3B} & \underline{74.96} & \underline{67.48} & 70.89 & 72.70 & 78.00 & 81.86 & 74.46 & \underline{75.62} & 47.72 & 65.46 & 69.21 & \underline{72.62} \\
\textbf{EVisRAG-7B} & \textbf{77.28} & \textbf{69.73} & \textbf{79.94} & \textbf{81.09} & \textbf{85.45} & \textbf{88.23} & \textbf{82.73} & \textbf{84.52} & \underline{52.10} & \textbf{67.67} & \textbf{75.50} & \textbf{78.25} \\
    \bottomrule
   \end{tabular}
    \label{tab:overall}
\end{table*}

\begin{table*}[t!]
    \centering
    \small
    \setlength{\tabcolsep}{9pt}
    \renewcommand{\arraystretch}{1.0}
    \caption{Ablation study on accuracy (\%) averaged over 5 runs with different random seeds. We report mean $\pm$ standard deviation. We compare \textit{Generic CoT} and \textit{Evidence-Guided Reasoning}, and further study three optimization strategies under evidence-guided reasoning: \textit{DAPO}, \textit{StepGRPO}, and \textit{RS-GRPO}.}
    \begin{tabular}{lccccc|c}
    \toprule
    \multirow{2}{*}{\textbf{Training Strategies}} &
    \multicolumn{2}{c}{\textbf{In Distribution}} &
    \multicolumn{3}{c}{\textbf{Out of Distribution}} &
    \multirow{2}{*}{\textbf{Avg. Acc}} \\
    \cmidrule(lr){2-3} \cmidrule(lr){4-6}
    & \textbf{ChartQA} & \textbf{InfoVQA} & \textbf{DocVQA} & \textbf{SlideVQA} & \textbf{ViDoSeek} & \\
    \midrule
    \textbf{Evidence-Guided Reasoning + RS-GRPO (Ours)} & $\textbf{76.8} \pm 0.6$ & $\textbf{79.2} \pm 0.7$ & $\textbf{85.5} \pm 1.2$ & $\textbf{81.3} \pm 1.0$ & $\textbf{51.8} \pm 0.7$ & $\textbf{74.9} \pm 0.8$ \\
    \midrule
    \hspace{1mm}Generic CoT + DAPO                     & $67.2 \pm 1.3$ & $73.3 \pm 1.6$ & $75.7 \pm 2.1$ & $77.3 \pm 2.0$ & $41.8 \pm 1.2$ & $67.1 \pm 1.6$ \\
    \hspace{1mm}Evidence-Guided Reasoning + DAPO      & $69.8 \pm 1.1$ & $74.2 \pm 1.2$ & $79.9 \pm 3.5$ & $77.5 \pm 2.2$ & $48.1 \pm 1.7$ & $69.9 \pm 1.9$ \\
    \hspace{1mm}Evidence-Guided Reasoning + StepGRPO  & $72.0 \pm 2.2$ & $75.7 \pm 2.1$ & $80.0 \pm 2.2$ & $77.9 \pm 1.7$ & $48.7 \pm 1.9$ & $70.9 \pm 2.0$ \\
    \bottomrule
    \end{tabular}
    \label{tab:ablation}
\end{table*}

\section{Results and Analysis}


\subsection{Overall Performance}

Table~\ref{tab:overall} reports the overall results for EVisRAG and all baselines. EVisRAG-7B consistently outperforms every comparator across all benchmarks, with substantial gains over the Qwen2.5-VL-7B backbone, averaging +19\% in accuracy and +27\% in F1 score. These improvements indicate that an evidence-guided reasoning paradigm, coupled with RS-GRPO, strengthens perceptual grounding and enables reasoning that is explicitly conditioned on grounded evidence. 
Compared with RL-trained VLRMs, EVisRAG’s explicit visual perception yields a clear advantage. Within the VLRM group, models emphasizing logical reasoning (e.g., OpenVLThinker~\citep{deng2025openvlthinker}) do improve question answering performance, underscoring the value of stronger reasoning. Nevertheless, EVisRAG’s added perceptual grounding closes a further gap.
Moreover, the three VRAG models improve the extraction of key evidence from retrieved context and, owing to their generalization, outperform the backbone when reasoning over multiple images. Yet they remain more than ten percentage points below EVisRAG-7B, since they neglect the need for strong perceptual grounding over multiple images with rich visual information. EVisRAG further allows a 7B parameter model to exceed the performance of considerably larger 32B parameter models. 
Furthermore, thrived on our RS-GRPO algorithm, EVisRAG jointly improves both perception and reasoning
capabilities of VLMs, leading to a more effective and adaptable RAG framework.


\begin{figure*}[!t]
  \centering
  \begin{subfigure}[b]{0.35\textwidth}
  \vspace{-0.2in}
    \includegraphics[width=\linewidth]{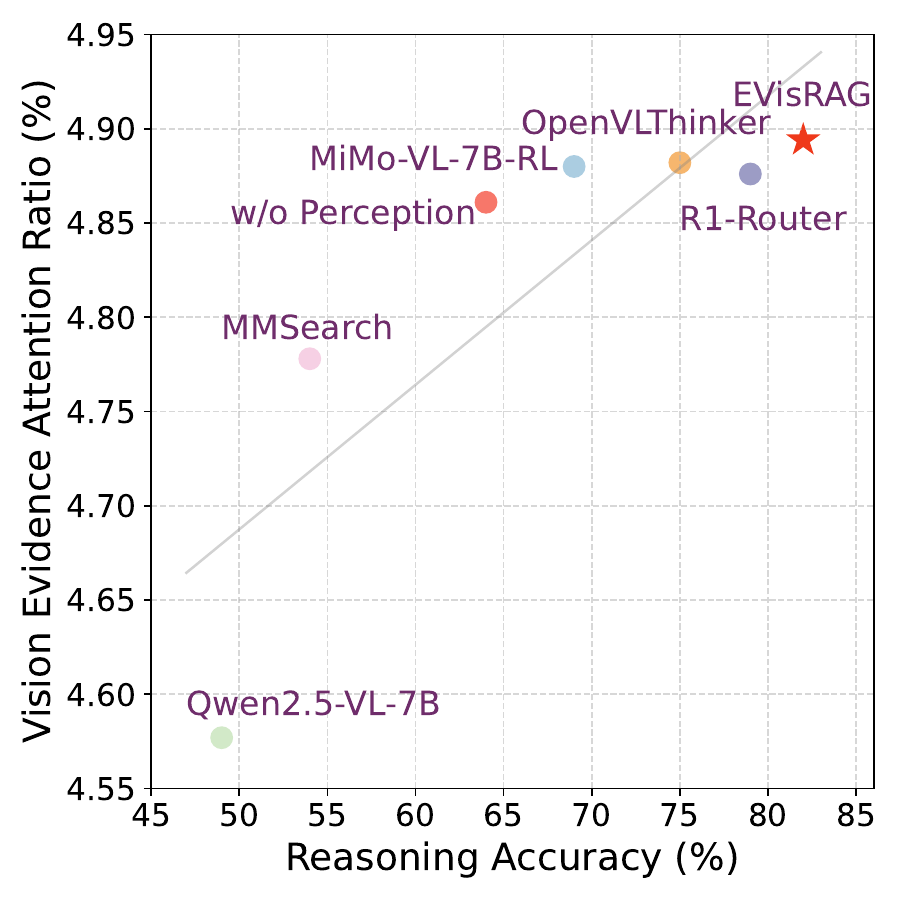}
    \caption{}
    \label{fig:perception_quantify}
  \end{subfigure}
  \begin{subfigure}[b]{0.62\textwidth}
    \includegraphics[width=\linewidth]{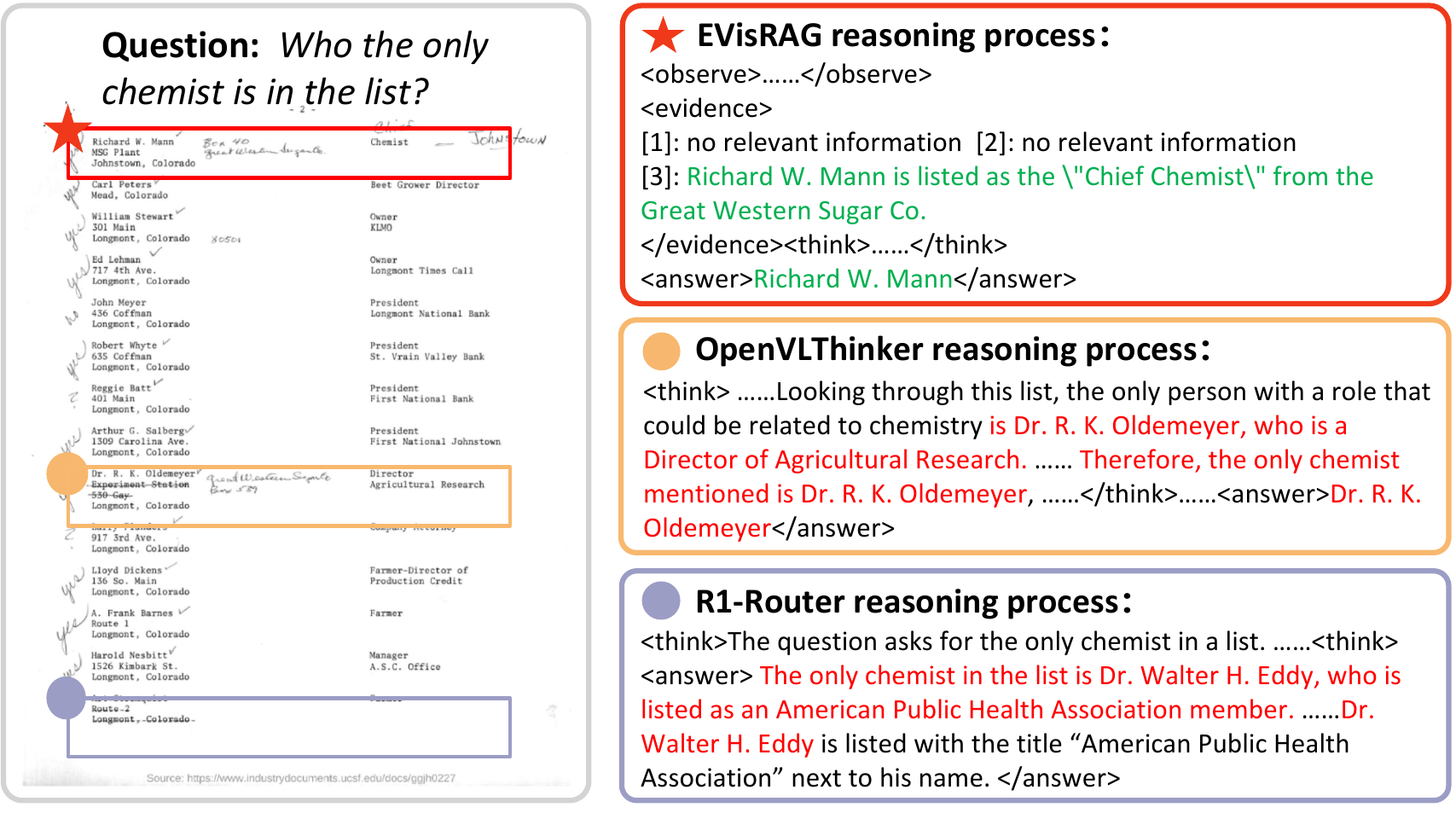}
    \caption{}
    \label{fig:perception_case}
  \end{subfigure}
\caption{Comparison of models' perception of question-relevant visual evidence. (a) Reasoning accuracy versus visual evidence attention ratio within human-annotated evidence boxes. EVisRAG achieves the highest reasoning accuracy and the highest attention ratio among all compared methods. (b) Case study on generated evidence and final answers. EVisRAG correctly identifies the question-relevant evidence that Richard W. Mann is the ``Chief Chemist'' and produces the correct answer.}
  \label{fig:perception}
\end{figure*}
\begin{figure*}[t!]
  \centering
  \includegraphics[width=1.0\textwidth]{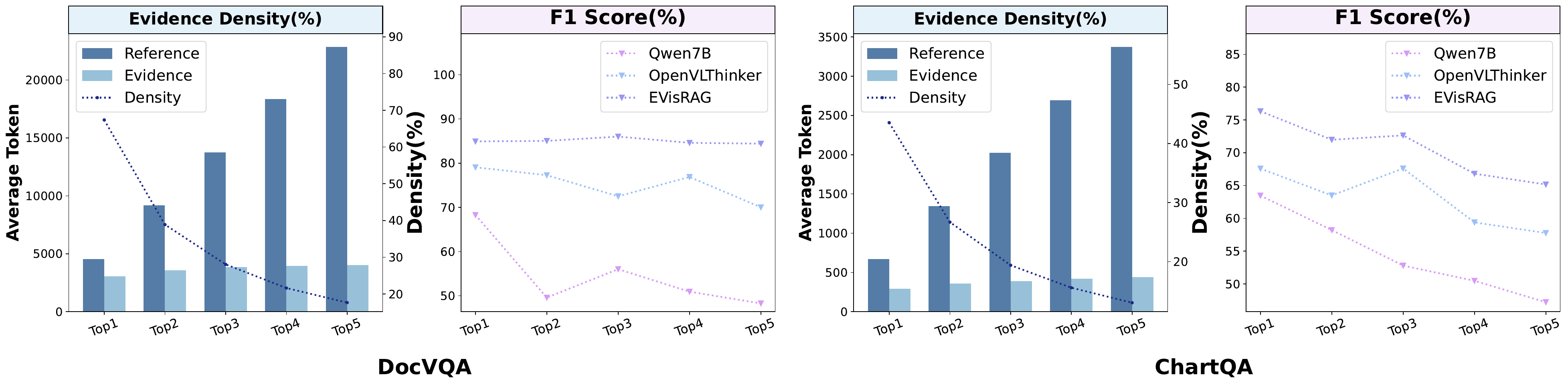}  
  \caption{Performance comparison on different visual evidence density. Despite increasing noise with more retrieved images, EVisRAG maintains stable.}
  \label{fig:density}
\end{figure*}

\subsection{Ablation Study}

This section studies EVisRAG as a progression of training strategies. As shown in Table~\ref{tab:ablation}, we compare a generic chain-of-thought baseline (\textit{Generic CoT + DAPO}) with three evidence-guided variants trained with different optimization strategies: \textit{DAPO}, \textit{StepGRPO}, and \textit{RS-GRPO}.

First, replacing \textit{Generic CoT + DAPO} with \textit{Evidence-Guided Reasoning + DAPO} yields consistent gains across all benchmarks. This verifies that the improvement does not come merely from generating longer reasoning traces, but from explicitly structuring the generation process around evidence localization and grounded reasoning. Second, under the same evidence-guided reasoning framework, performance improves steadily from \textit{DAPO} to \textit{StepGRPO} and further to \textit{RS-GRPO}. The comparison between \textit{DAPO} and \textit{StepGRPO} shows that perception reward provides useful supervision for intermediate visual grounding. The comparison between \textit{StepGRPO} and \textit{RS-GRPO} further shows that the key advantage of RS-GRPO lies in scoped credit assignment: instead of broadcasting all rewards to the entire sequence, RS-GRPO restricts each reward to the corresponding token span. This design reduces cross-stage interference, sharpens optimization signals, and ultimately delivers the best overall results.

\subsection{Evaluating Perceptual Ability through Attention and Evidence Quality}
In this section, we evaluate EVisRAG’s perceptual ability from both quantitative and qualitative perspectives. Figure~\ref{fig:perception_quantify} reports the visual evidence attention ratio, defined as the percentage of attention mass falling inside human-annotated evidence boxes. As shown in the figure, EVisRAG lies in the top-right region among all compared methods, achieving both the highest reasoning accuracy and the highest evidence attention ratio. This result suggests that EVisRAG is better at concentrating on question-relevant visual regions, and that such evidence-focused attention is closely associated with stronger downstream reasoning performance.

Figure~\ref{fig:perception_case} further provides a case study on generated evidence and final answers. For the question ``Who the only chemist is in the list?'', EVisRAG first filters out irrelevant pages by explicitly recording ``no relevant information'' for unrelated entries, and then correctly identifies the supporting evidence that Richard W. Mann is listed as the ``Chief Chemist.'' Based on this evidence, it produces the correct final answer. In contrast, OpenVLThinker~\cite{deng2025openvlthinker} incorrectly treats a chemistry-related but irrelevant title (``Director of Agricultural Research'') as supporting evidence, while R1-Router~\cite{peng2025learning} outputs an answer that is not grounded in the relevant entry at all. 

Together, these results show that EVisRAG not only attends more to question-relevant regions but also transforms such perception into cleaner and more faithful evidence records. This stronger evidence quality is crucial for grounded multi-image reasoning, as it reduces distraction from irrelevant visual details and leads to more reliable final predictions.

\subsection{Visual Evidence Density Comparison}

We perform a visual evidence density analysis to evaluate the robustness of our method under varying numbers of images and different evidence densities. As shown in Figure~\ref{fig:density}, for each question, we retrieve the top-1 to top-5 images as context, which we refer to as references. Within these references, image tokens that provide information supporting the answer are defined as Evidence. It can be observed that as the number of retrieved images increases, the total number of Evidence tokens also rises. However, the overall evidence density decreases rapidly. We compare the performance of our method with Qwen-7B (backbone) and OpenVLThinker (the strongest baseline) in terms of F1 score across different evidence densities. Our method consistently outperforms both baselines at all density levels, demonstrating its superior generalizability in multi-image scenarios. Furthermore, on the DocVQA dataset, our approach maintains stable performance even as the evidence density decreases, highlighting its strong ability to resist hallucination effects.



\begin{figure*}[t!]
  \centering
  \includegraphics[width=0.9\textwidth]{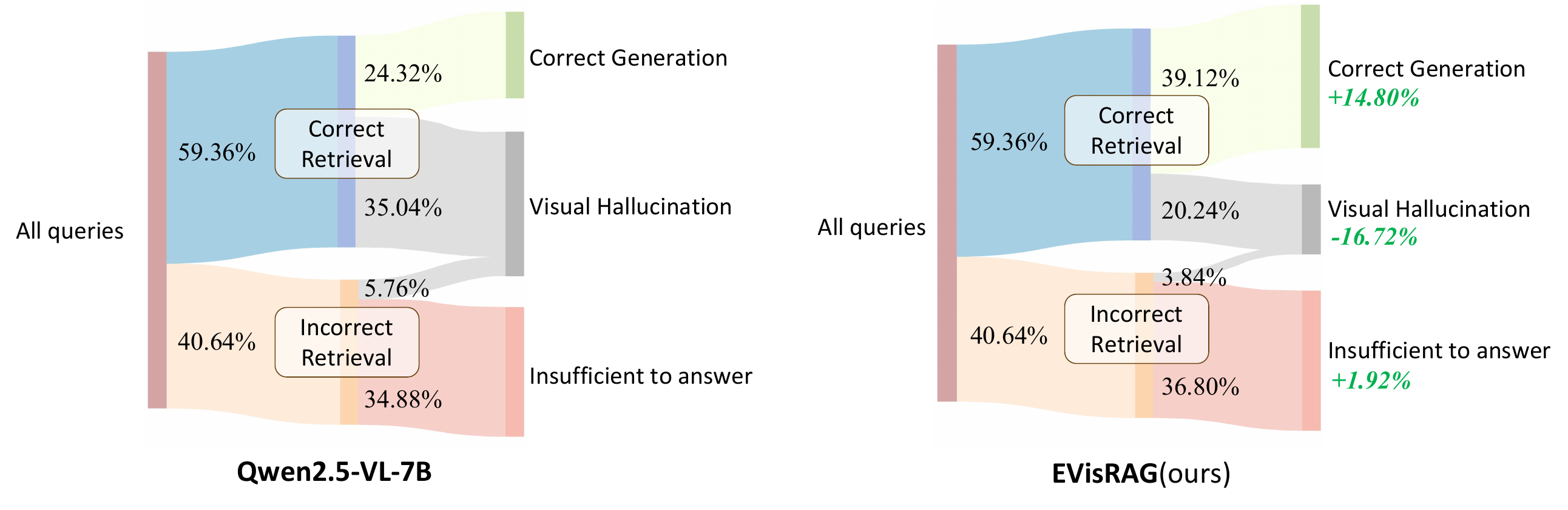}  
  \caption{Model performance comparisons in different retrieval scenarios on ChartQA. Compared with the backbone, EVisRAG remains more faithful to the retrieved content in both correct and incorrect retrieval scenarios.}
  \label{fig:retrieval}
\end{figure*}

\begin{figure}[!t]
  \centering
    \includegraphics[width=\linewidth]{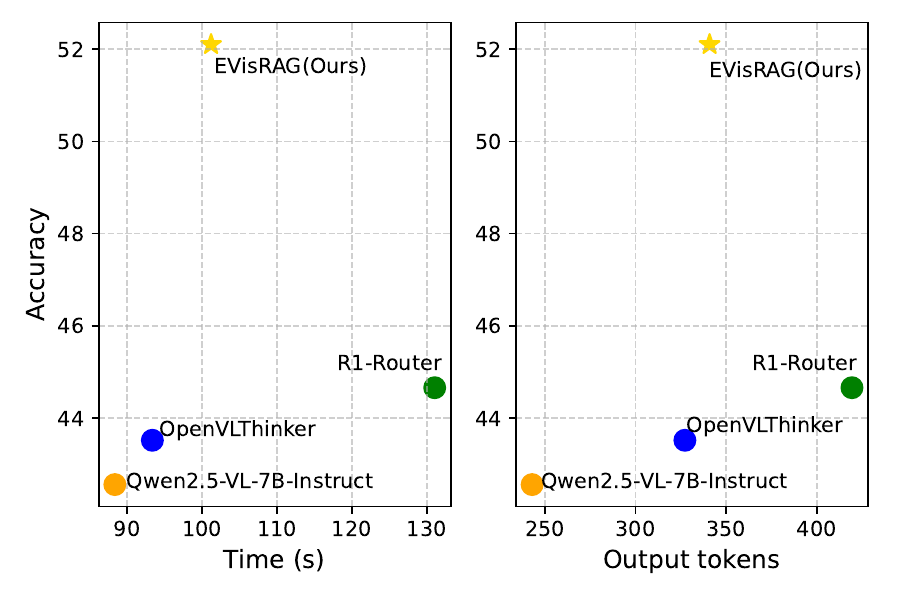}
    \caption{Inference Efficiency Comparison of EVisRAG}
  \label{fig:inference}
\end{figure}
 
\subsection{Impact of Training on Model Performance}
As shown in Figure~\ref{fig:retrieval}, we examine the model’s reasoning under varying degrees of contextual sufficiency to evaluate its balance between informativeness and hallucination. Before training, the backbone model displays a strong hallucination tendency even with correct retrieval—only 24.32\% of queries yield correct generations, whereas 35.04\% produce incorrect responses. Under incorrect retrieval, it predominantly abstains, reflecting a conservative strategy typical of smaller models. After training with our method, EVisRAG achieves a markedly better performance: the correct-generation rate in correctly retrieved contexts rises substantially, while a modest increase in abstention under incorrect retrieval is accompanied by a controlled reduction in incorrect generations. Overall, EVisRAG exhibits strengthened evidence-sensitive reasoning and reduced visual hallucination in underdetermined scenarios.


\subsection{Inference Efficiency of EVisRAG}
Figure~\ref{fig:inference} compares inference accuracy, latency, and output length on the ViDoSeek dataset across different approaches. Baseline models such as Qwen2.5-VL-7B-Instruct and OpenVLThinker exhibit relatively low inference time (around 90–95 seconds) and short outputs (approximately 270–330 tokens), but their accuracy remains below 44\%. R1-Router does not improve accuracy while incurring the highest computational cost: it requires the longest inference time (about 120 seconds) and produces the most verbose outputs (over 400 tokens) for a similar accuracy level. In contrast, our proposed EVisRAG, which adopts a single-step generation strategy, achieves a substantial accuracy gain of over 52\% with only a modest increase in latency (around 100 seconds) and a moderate number of output tokens (about 300). These results show that EVisRAG delivers significantly better reasoning quality without sacrificing efficiency or introducing excessive verbosity, demonstrating its practicality for real-world applications.



\subsection{Comparison with MCOT Prompting Strategies}

To evaluate the effectiveness of Evidence-Guided Reasoning, which explicitly encourages the VLM to first observe and record visual evidence before reasoning, we conduct two additional experiments. First, we compare our reasoning paradigm with three MCOT baselines, which also require no additional training and instead attempt to improve multimodal reasoning through fixed prompting patterns. As shown in Table~\ref{tab:mcot}, our approach consistently outperforms all MCOT baselines across all five datasets, demonstrating that the advantage of Evidence-Guided Reasoning is not limited to a particular benchmark but generalizes across diverse visual question answering settings. In particular, our method achieves the best average performance by a clear margin, indicating that explicitly structuring the reasoning process around evidence collection is more effective than relying on handcrafted prompting templates alone.


\begin{table}[t]
   \centering
      \caption{Overall Performance of EvidenceCOT and Other MCOT.}
      \setlength{\tabcolsep}{3pt}
   \begin{tabular}{lcccccc} \toprule 
   \textbf{Methods} &\textbf{C-QA} & \textbf{I-VQA} & \textbf{D-VQA} & \textbf{S-VQA} & \textbf{VDS} & \textbf{Avg.} \\
 \midrule
      COCOT~\cite{zhang2024cocot} & 49.52 & 31.34 & 40.95 & 35.25& 33.80& 38.17 \\ 
      CCOT~\cite{mitra2024compositional} & 50.32& 36.91 & 41.96 & 51.80 & 36.16 & 43.43 \\ 
      DDCOT~\cite{zheng2023ddcot} & 51.68 & 43.73 & 62.10& 54.14 & 42.21 & 50.77  \\  \midrule
      EvidenceCOT & \textbf{62.72}& \textbf{65.94} & \textbf{70.05} & \textbf{66.73} & \textbf{46.50} & \textbf{62.39}  \\ \midrule
   \end{tabular}
   \label{tab:mcot}
\end{table}

\section{Conclusion}

In this paper, we propose EVisRAG,  a visual retrieval-augmented reasoning framework that enables Vision--Language Models (VLMs) to observe and localize evidence across multiple images during reasoning, thereby improving visual grounding and mitigating visual hallucinations in complex multi-image scenarios. To support this process, we introduce Reward-Scoped Group Relative Policy Optimization (RS-GRPO), which assigns reward credits to scope-specific token spans, stabilizing long chain-of-thought (CoT) training and encouraging more accurate evidence grounding. Extensive experiments demonstrate that EVisRAG can effectively aggregate key visual evidence, leading to substantial gains in reasoning accuracy while reducing hallucinated responses. Further analysis shows that its improvements come not simply from generating more descriptions, but from identifying question-relevant evidence and using it to support more reliable answer derivation. Overall, EVisRAG represents a promising step toward more reliable, interpretable, and hallucination-resistant visual retrieval-augmented generation systems.


\bibliographystyle{ACM-Reference-Format}
\bibliography{sample-base}

\appendix
\clearpage
\section{Appendix}


\subsection{Datasets}
\label{app:datasets}

We evaluated five VQA benchmarks: InfoVQA, DocVQA, and SlideVQA were obtained from the VisRAG release~\citep{yu2024visrag}, ChartQA from its test split~\citep{masry2022chartqa}, and ViDoSeek from ViDoRAG~\citep{wang2025vidorag}. Each dataset provides ground-truth answer image IDs. For each question, we retrieved the top-3 images using VisRAG-Ret as contexts. Following \citet{joren2024sufficient}, we labeled it sufficient if all ground-truth images were included, otherwise insufficient. The number of questions and the sufficient context ratio in the dataset are shown in Table~\ref{tab:datasets}.

\begin{table*}[htbp]
\centering
\small \setlength{\tabcolsep}{18pt}
\caption{Datasets used in our experiments.}
\begin{tabular}{lccc}
\toprule
\textbf{Name} & \textbf{\#Questions} & \textbf{Description} & \textbf{Sufficient Context Ratio} \\
\midrule
ChartQA & 1250& Visual and Logical Reasoning about Charts & 59.36\% \\
InfoVQA & 718 & Question Answering on Infographic Images & 92.90\% \\
DocVQA & 591 & Document Visual Question Answering & 83.59\% \\
SlideVQA & 556 & Question Answering based on Multiple Slides & 89.93\% \\
ViDoSeek & 1142 & Retrieval and Reasoning on Visually Rich Documents & 84.24\% \\
\bottomrule
\end{tabular}
\label{tab:datasets}
\end{table*}

\subsection{Data Construction of Golden Reasoning Trajectories}
\label{app:data_construction}

For model training, we collected 30,000 samples from the ChartQA~\citep{masry2022chartqa} and InfographicsVQA~\citep{mathew2022infographicvqa} datasets, which were randomly divided into two subsets for SFT and GRPO in an 8:2 ratio. During the retrieval stage, VisRAG-Ret retrieves the top five candidate images for each query. While evaluation uses only the top three images as context, training leverages a variable number of retrieved images (top-1 to top-5) for data augmentation. Reasoning trajectories are constructed by generating candidate chains of thought with Qwen2.5-VL-72B and Qwen2.5-VL-7B~\citep{bai2025qwen2}, followed by a filtering process that retains only those trajectories yielding correct answers. This procedure generated 60,000 high-quality samples for SFT training, from which we extracted evidence as Ground Truth Evidence.

\begin{figure*}[htbp]
  \centering
  \includegraphics[width=0.8\textwidth]{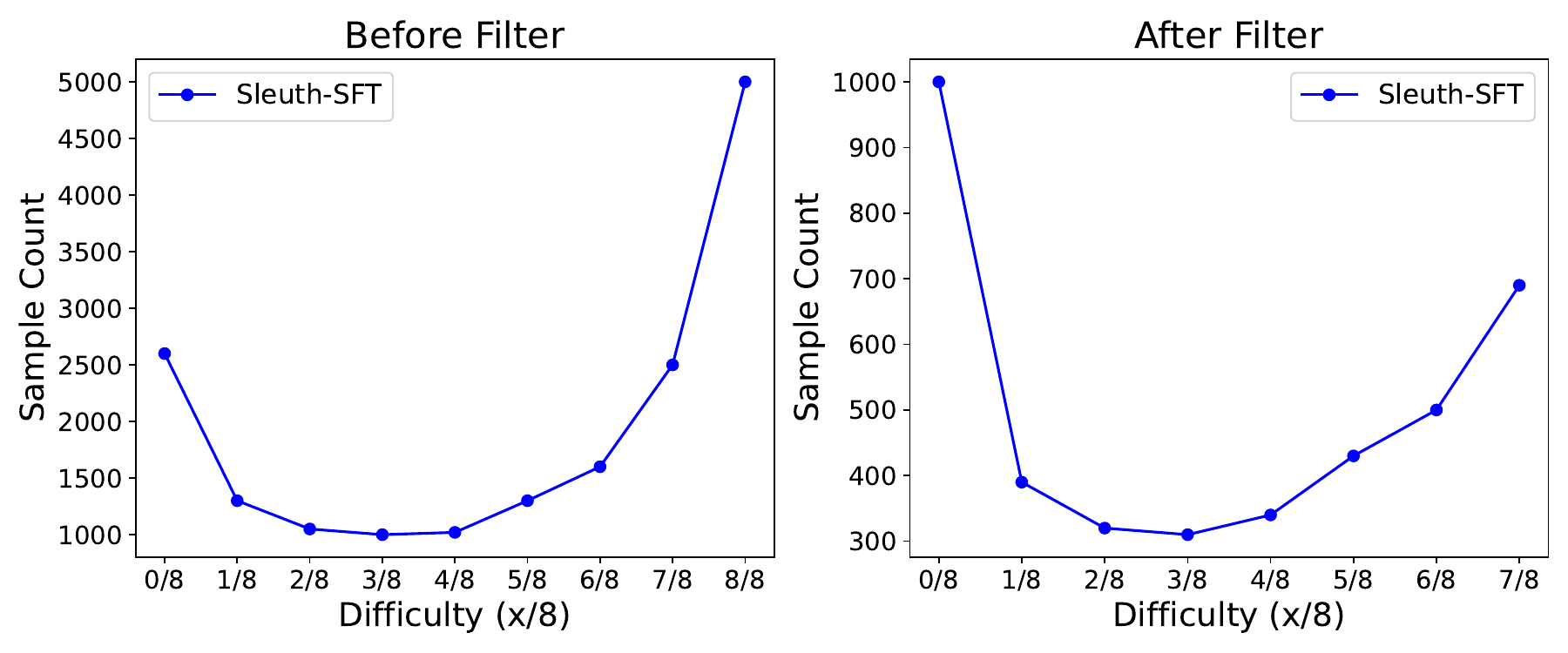}  
  \caption{Data Difficulty Distribution of Before-Filtering and After-Filtering.}
  \label{fig:dataconstruct}
\end{figure*}

In the GRPO phase, we adopt a curriculum learning strategy, following~\citep{Polaris2025}. Specifically, the SFT-trained model generates eight candidate completions for each sample, which are ranked according to their scores. Completions with perfect scores are excluded to mitigate overfitting. In addition, we incorporate 400 more challenging multi-hop examples from MMLongBench~\citep{ma2024mmlongbench}. The final GRPO training set consists of 4,000 carefully curated samples, organized to ensure a smooth progression from simple to complex instances, with a deliberate emphasis on more difficult cases to strengthen the model’s reasoning robustness. The distributions of data difficulty before and after filtering are illustrated in Figure~\ref{fig:dataconstruct}.

\subsection{More Details on the Fine-grained Reward}
\label{app:reward}
In addition to stabilizing training to ensure accurate perceptual grounding and evidence-guided reasoning in VLMs, we further introduce five reward components, namely:

\textbf{Perception reward.} For text-only language models, using only answer accuracy as the reward signal together with GRPO training can elicit emergent ``aha moments" and strengthen reasoning abilities~\citep{guo2025deepseek}. For VLMs, however, directly optimizing answer accuracy may improve reasoning while failing to improve perceptual accuracy~\citep{liu2025more}. To optimize perceptual grounding and reasoning at the same time during training, we introduce a fine-grained perception reward:

\begin{equation}
{\small
\begin{gathered}
R_{\text{perception}} = 
\frac{\sum_{i=1}^n r_i}{\sum_{i=1}^n \left( y_i \cdot k_{\text{pos}} + (1 - y_i) \cdot 1 \right)}, \\
r_i =
\begin{cases}
k_{\text{pos}} \cdot f_1\!\left(e_i^{\text{pred}}, e_i^{\text{gold}}\right), & \text{if } y_i = 1, \\
\mathbb{I}\!\left(e_i^{\text{pred}} = \text{``no relevant information''}\right), & \text{if } y_i = 0.
\end{cases}
\end{gathered}
}
\label{eq:percreward}
\end{equation}

The perception reward assesses whether the model extracts useful visual information. For each image, the evidence recorded by the model is compared with the corresponding gold evidence. For images that contain information relevant to the question, the reward is the F1 score between the predicted and gold evidence. For images that are irrelevant, the reward equals 1 if the model correctly indicates the absence of evidence and 0 otherwise. The final perception reward is the normalized sum of the image level rewards. 

\textbf{Derivation reward.} We employ the F1-score between the predicted answer and the gold truth as the reasoning reward, where the gold truth is set to the fixed response ``insufficient to answer" when the context is incomplete.

\begin{equation}
{\small
\begin{gathered}
R_{\text{derivation}} = f_1(a^{\text{pred}}, a^{\text{gold}}), \\
a^{\text{gold}} = 
\begin{cases}
a^{\text{gold}}, & \text{if sufficient context} \\
\text{``insufficient to answer"}, & \text{if insufficient context}
\end{cases}
\end{gathered}
}
\label{eq:reward}
\end{equation}

where $a^{\text{pred}}$ denotes the model’s predicted answer, $a^{\text{gold}}$ denotes the ground-truth answer, and $\text{Acc}_{\text{evi}}$ indicates whether the model's evidence predictions for all images are correct (assigned $1$ if all are correct, and $0$ otherwise).

\textbf{Format reward.} Beyond the accuracy-based reward, we also incorporate a format reward model that compels the model to follow our CoT design by sequentially performing observation, evidence recording, reasoning, and answering, with each stage encapsulated by its corresponding special tag (\textless observe\textgreater, \textless evidence\textgreater, \textless think\textgreater, \textless answer\textgreater).
\begin{equation}
\small
R_{\text{format}}(a_i) =
\begin{cases}
1, & \text{if the format of $a_i$ is correct} \\
0, & \text{otherwise}
\end{cases}
\label{eq:reward2}
\end{equation}

\subsection{Impact of Evidence-Guided Reasoning}
\label{app:abcot}

To further validate the generality and robustness of our evidence-guided reasoning paradigm, we examine it along two complementary dimensions: transferability across model architectures and consistency across model scales.

To assess whether the benefits of evidence-guided reasoning extend beyond the Qwen-VL family, we apply the architecture-agnostic EVisRAG-Prompt to InternVL3-8B. As shown in the table below, this improves the average accuracy from 57.72 to 60.50 and yields gains on 4 out of 5 datasets. This indicates that the EVisRAG paradigm can transfer beyond the Qwen-VL family, although full RL-based transfer to other VLM architectures remains future work.

\begin{table}[t]
\centering
\caption{Transferability of EVisRAG to InternVL3-8B.}
\label{tab:internvl_transfer}
\small
\setlength{\tabcolsep}{4pt}
\renewcommand{\arraystretch}{0.9}
\begin{tabular}{lccccc}
\toprule
\textbf{Method} & \textbf{C-QA} & \textbf{I-VQA} & \textbf{D-VQA} & \textbf{S-VQA} & \textbf{VDS} \\
\midrule
InternVL3-8B 
& 55.44 & 59.47 & 67.01 & 63.49 & 43.17 \\
\quad w/ EVisRAG-Prompt
& 61.12 & 55.29 & 73.60 & 64.57 & 47.90 \\
\bottomrule
\end{tabular}
\vspace{-14pt}
\end{table}

\begin{figure*}[t!]
  \centering
  \includegraphics[width=1.0\textwidth]{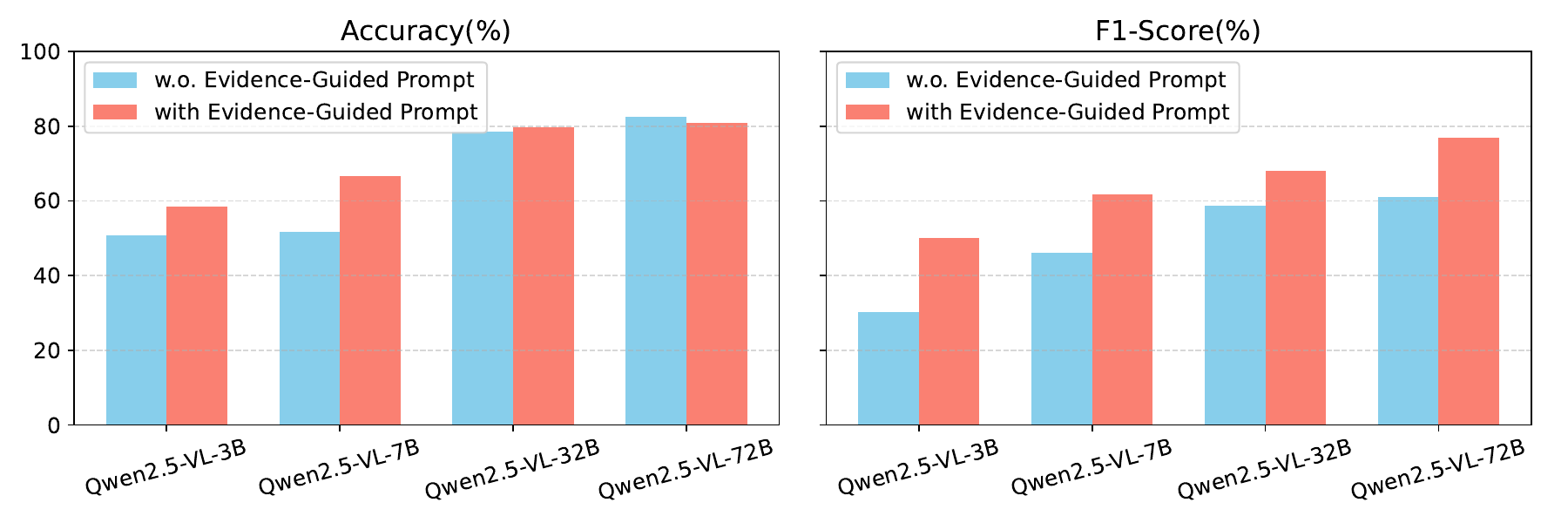}  
  \caption{Performance comparison of Evidence-Guided Prompt Approach across different model sizes on the SlideVQA dataset.}
  \label{fig:ab_cot}
\end{figure*}

We further evaluate whether Evidence-Guided Prompting remains effective across models of different scales. As illustrated in Figure~\ref{fig:ab_cot}, even without additional training, prompting the model to first record visual evidence and then reason upon it consistently improves both perception and reasoning across four different model sizes. Together with the cross-architecture results above, this demonstrates the broad applicability and robustness of our proposed paradigm across both architectures and scales. Prompt templates used by EVisRAG are shown in Figure~\ref{fig:evidencetrain_prompt}.

\subsection{More Implementation Details}
\label{app:trainparameters}
We acknowledge the contributions of LLaMA-Factory and EasyR1~\citep{zheng2025easyr1} for releasing the training frameworks utilized in our SFT and GRPO experiments. We adopt Qwen2.5-VL-7B~\citep{bai2025qwen2} as the backbone model for our proposed EVisRAG. EVisRAG is trained on 8× NVIDIA A100-80GB GPUs, with hyperparameters as shown in Tables~\ref{tab:sft_hyperparams} and~\ref{tab:grpo_hyperparams}.

\begin{table*}[htbp]
  \centering
  \begin{minipage}{0.48\textwidth}
    \centering
    \caption{SFT hyperparameters.}
    \begin{tabular}{lc}
      \toprule
      Epoch & 1 \\
      Data type & bf16 \\
      Learning rate & 5e-7 \\
      Global batch size & 32 \\
      Scheduler & Cosine \\
      Warmup ratio & 0.1 \\
      Num train epochs & 1 \\
      Image max pixels & 3920000 \\
      \bottomrule
    \end{tabular}
    \label{tab:sft_hyperparams}
  \end{minipage}
  \hfill
  \begin{minipage}{0.48\textwidth}
    \centering
    \caption{GRPO hyperparameters.}
    \begin{tabular}{lc}
      \toprule
      Epoch & 4 \\
      Rollout batch size & 32 \\
      Global batch size & 32 \\
      Max grad norm & 1.0 \\
      Data type & bf16 \\
      Learning rate & 1e-6 \\
      Weight decay & 1e-2 \\
      Warmup ratio & 0.0 \\
      Rollout temperature & 1.2 \\
      epsilon & 0.2 \\
      epsilon\_high & 0.28 \\
      Image max pixels & 1568000 \\
      \bottomrule
    \end{tabular}
    \label{tab:grpo_hyperparams}
  \end{minipage}
\end{table*}

\subsection{More Implementation Details of the Baseline Methods}
\label{app:baselines}

In this section, we provide comprehensive implementation details and prompt templates of the baseline methods evaluated in our study.

\textbf{General VLMs.} We assessed general vision-language models across different scales, namely Qwen2.5-VL-7B and Qwen2.5-VL-32B~\citep{bai2025qwen2}, as well as MiMo-VL-7B-RL~\citep{coreteam2025mimovltechnicalreport}.

\textbf{VisRAG-Gen.} We additionally evaluate two generation strategies described in VisRAG~\citep{yu2024visrag}.

\textit{Page Concatenation.} Page Concatenation forms a single composite image by horizontally concatenating the top-$k$ retrieved pages and feeds it to a single-image VLM. In our implementation, we adopt Qwen2.5-VL-7B~\citep{bai2025qwen2} as the backbone VLM to ensure a fair comparison with other strong VRAG systems.

\textit{Weighted Selection.} Weighted Selection instead generates an answer for each retrieved page independently and selects the final output based on the highest confidence, where the confidence weight combines the generation likelihood and the normalized retrieval score. For this method, we use the official implementation and pretrained MiniCPM-V-2~\citep{yao2024minicpmvgpt4vlevelmllm} model released by the authors. Together, these two variants represent the canonical generation pipelines of VisRAG and serve as competitive baselines in our evaluation.

\textbf{Vision-Language Reasoning Models (VLRMs).} We compare five fine-tuned VLRMs, all initialized from Qwen2.5-VL-7B-Instruct, each employing distinct strategies to enhance reasoning capabilities:

\textit{Vision-R1-7B.} Vision-R1-7B~\citep{zhan2025vision} introduces a reinforcement learning–based fine-tuning approach that incentivizes reasoning through vision-guided feedback. It circumvents the need for human-curated preference data by adopting a criterion-driven reward function.

\textit{OpenVLThinker-7B.} OpenVLThinker-7B~\citep{deng2025openvlthinker} follows an iterative two-stage training scheme, alternating between supervised fine-tuning (SFT) and reinforcement learning (RL). Starting from distilled reasoning competencies in text-only domains, the model progressively refines its reasoning by generating its own training data through RL and then using that data to further supervise and fine-tune itself.

\textit{MM-Eureka-7B.} MM-Eureka-7B~\citep{meng2025mm} extends rule-based reinforcement learning (RL) to multimodal reasoning by incorporating new algorithms such as Online Filter, ADORA, and DAPO, which enhance reasoning efficiency and stability across multimodal tasks.

\textit{Ocean-R1-7B.} Ocean-R1-7B~\citep{ming2025oceanr1} builds upon a structured chain-of-thought evaluation framework that leverages knowledge graph exploration (e.g., OCEAN) to provide rich offline feedback, thereby aligning generated reasoning paths with factual knowledge.

\textit{ThinkLite-VL-7B.} ThinkLite-VL-7B~\citep{wang2025sota} employs Monte Carlo Tree Search (MCTS)–guided sample selection to identify and train on genuinely challenging examples from a small dataset (11k samples), achieving state-of-the-art visual reasoning performance with high data efficiency.

\textbf{VRAGs (Visual Retrieval-Augmented Generation).} We further examine three advanced VRAG methods, all built upon the Qwen2.5-VL-7B-Instruct architecture:

\textit{R1-Router.} R1-Router~\citep{peng2025learning} employs a dynamic routing mechanism trained via Step-wise Group Relative Policy Optimization (Step-GRPO). R1-Router generates intermediate queries during the model’s reasoning process and directs them selectively to the most appropriate knowledge base (e.g., text, image, table KB), harnessing the evolving reasoning state. This fine-grained routing capability enhances retrieval efficiency and reasoning precision by minimizing unnecessary retrievals while adaptively integrating external evidence.

\textit{MMSearch-R1.} MMSearch-R1~\citep{wu2025mmsearch} integrates multimodal search into the reasoning loop, employing cross-modal retrieval mechanisms to fetch contextually aligned information in both visual and textual forms.

\textit{VRAG-RL.} VRAG-RL~\citep{wang2025vrag} incorporates a reinforcement learning–based fine-tuning schema, enabling the model to progressively gather visual evidence from coarse to fine granularity and support multi-turn reasoning via an optimized retrieval-and-generation pipeline.

\begin{table*}[t!]
   \centering
   \small \setlength{\tabcolsep}{8pt}
  \caption{Overall Performance of EVisRAG and VisRAG-Gen. \textbf{Bold} denotes the highest value.}
   \begin{tabular}{lcccccccccccc} \toprule 
   \multirow{3}{*}{\textbf{Methods}} & \multicolumn{4}{c}{\textbf{In Distribution}} & \multicolumn{6}{c}{\textbf{Out of Distribution}} & \multicolumn{2}{c}{\multirow{2}{*}{\textbf{Average}}} \\ \cmidrule(lr){2-5} \cmidrule(lr){6-11} 
      ~ & \multicolumn{2}{c}{\textbf{ChartQA}} & \multicolumn{2}{c}{\textbf{InfoVQA}} & \multicolumn{2}{c}{\textbf{DocVQA}} & \multicolumn{2}{c}{\textbf{SlideVQA}} & \multicolumn{2}{c}{\textbf{ViDoSeek
      }} & ~ & ~\\ \cmidrule(lr){2-5} \cmidrule(lr){6-11} \cmidrule(lr){12-13}
      ~ & Acc & F1 & Acc & F1 & Acc & F1 & Acc & F1 & Acc & F1 & Acc & F1 \\ \midrule
    \rowcolor{gray!20} \multicolumn{13}{l}{\textbf{VisRAG-Gen}} \\ \midrule
      Page Concatenation & 59.20 & 52.80 & 52.92 & 46.42 & 60.58 & 47.84 & 64.57 & 48.45 & 45.01 & 41.37 & 56.63 & 47.63 \\ 
      Weighted Selection & 32.24 & 32.32 & 25.07 & 27.36 & 33.67 & 37.11 & 33.81 & 36.44 & 21.98 & 31.64 & 29.35 & 32.97 \\ \midrule
      \rowcolor{gray!20} \multicolumn{13}{l}{\textbf{EVisRAG(ours)}} \\ \midrule
        \textbf{EVisRAG-3B} & 72.64 & 72.54 & 71.03 & 71.83 & 78.17 & 79.30 & 75.84 & 75.49 & 45.71 & 60.13 & 68.68 & 71.86 \\
        \textbf{EVisRAG-7B} & \textbf{76.80} & \textbf{76.60} & \textbf{79.39} & \textbf{79.80} & \textbf{85.45} & \textbf{86.82} & \textbf{81.29} & \textbf{80.28} & \textbf{52.10} & \textbf{65.78} & \textbf{75.01} & \textbf{77.86} \\ 
    \bottomrule
   \end{tabular}
   \label{tab:visraggen}
\end{table*}

\begin{figure*}[t!]
  \centering
  \includegraphics[width=\textwidth]{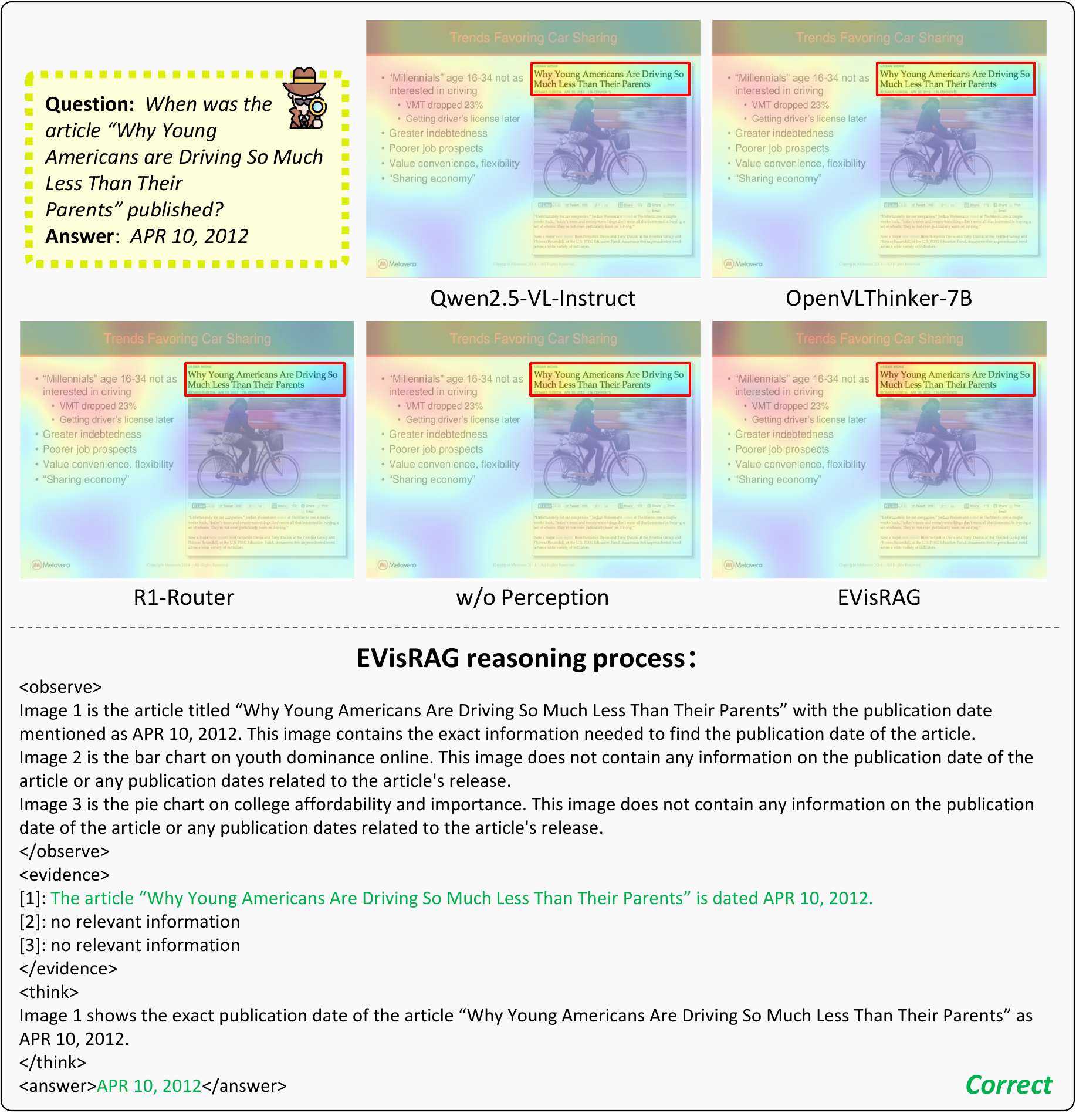}  
  \caption{A Evidence Attention Case Study on SlideVQA.}
  \label{fig:appd_attn_case}
\end{figure*}

\begin{figure*}[t!]
  \centering
  \includegraphics[width=0.8\textwidth]{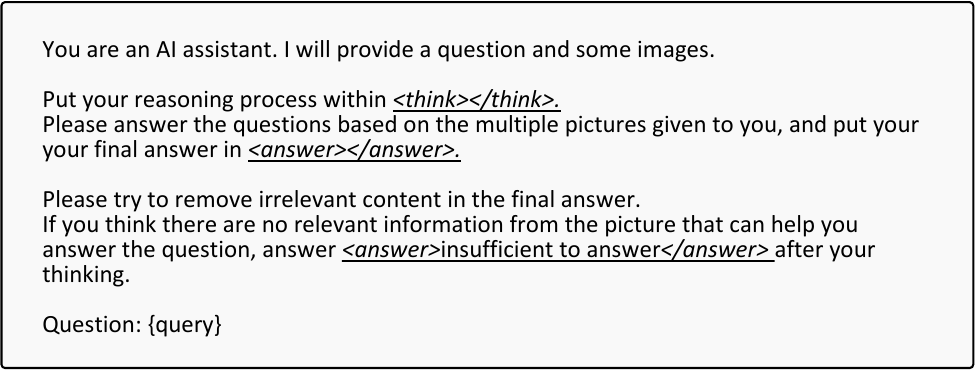}  
  \caption{The Prompt Template for baselines.}
  \label{fig:baseline_prompt}
\end{figure*}

\begin{figure*}[t!]
  \centering
  \includegraphics[width=0.8\textwidth]{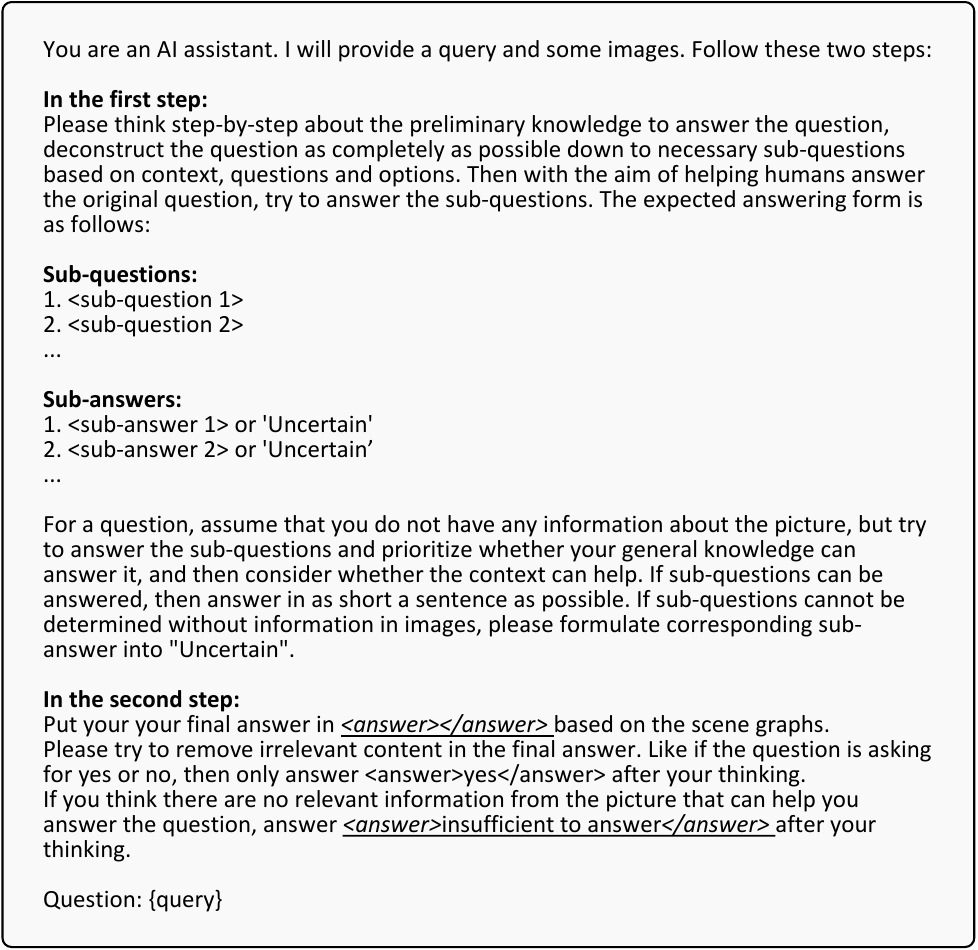}  
  \caption{The Prompt Template for DDCOT}
  \label{fig:ddcot_prompt}
\end{figure*}

\begin{figure*}[t!]
  \centering
  \includegraphics[width=0.8\textwidth]{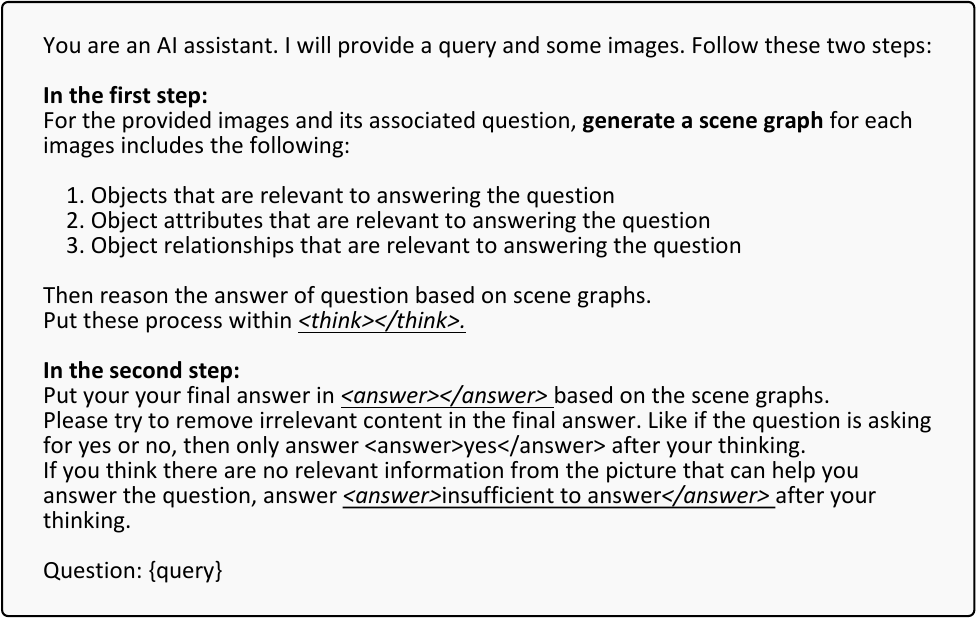}  
  \caption{The Prompt Template for CCOT}
  \label{fig:ccot_prompt}
\end{figure*}

\begin{figure*}[t!]
  \centering
  \includegraphics[width=0.8\textwidth]{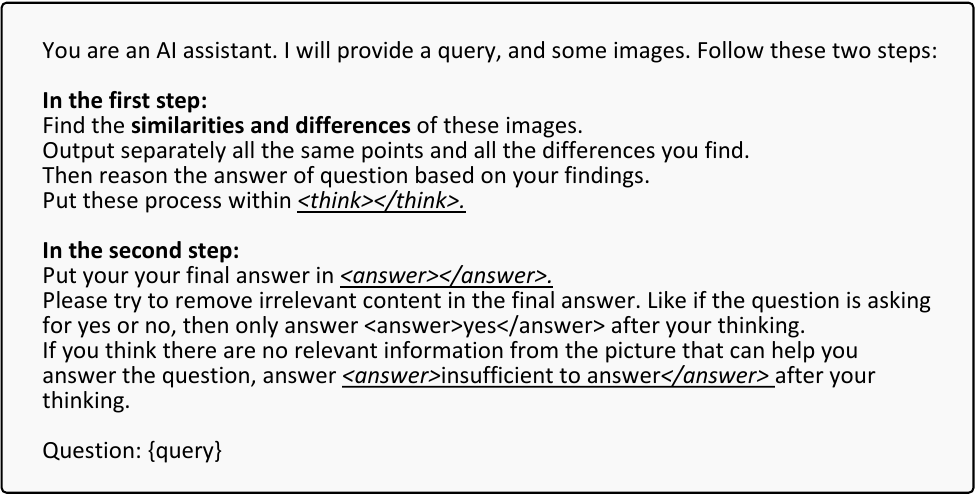}  
  \caption{The Prompt Template for COCOT}
  \label{fig:cocot_prompt}
\end{figure*}

The prompt templates employed for each baseline are shown in Figure~\ref{fig:baseline_prompt}. For the three MCOT-based comparisons (DDCOT, CCOT, and COCOT), we adapted their original prompting strategies into an end-to-end chain-of-thought generation framework compatible with our setup. Their corresponding prompt templates are detailed in Figures~\ref{fig:ddcot_prompt},~\ref{fig:ccot_prompt}, and~\ref{fig:cocot_prompt}, respectively.

\textbf{VisRAG-Gen.} We also compared two generation methods in VisRAG~\cite{yu2024visrag}, results shown in~\ref{tab:visraggen}:

\textit{Page Concatenation.} Page Concatenation concatenates all retrieved images into a single composite image, which is then fed to a vision–language model that only supports single-image inputs; we adopt Qwen2.5-7B-VL as the backbone model for this setting.

\textit{Weighted Selection.} Following VisRAG, we prompt the model to generate an answer for each retrieved image independently and select the final answer based on a confidence-weighted scoring scheme provided by the original implementation.

\subsection{Extended Ablation with Bootstrap Confidence Intervals}
\begin{figure*}[t!]
  \centering
  \includegraphics[width=\textwidth]{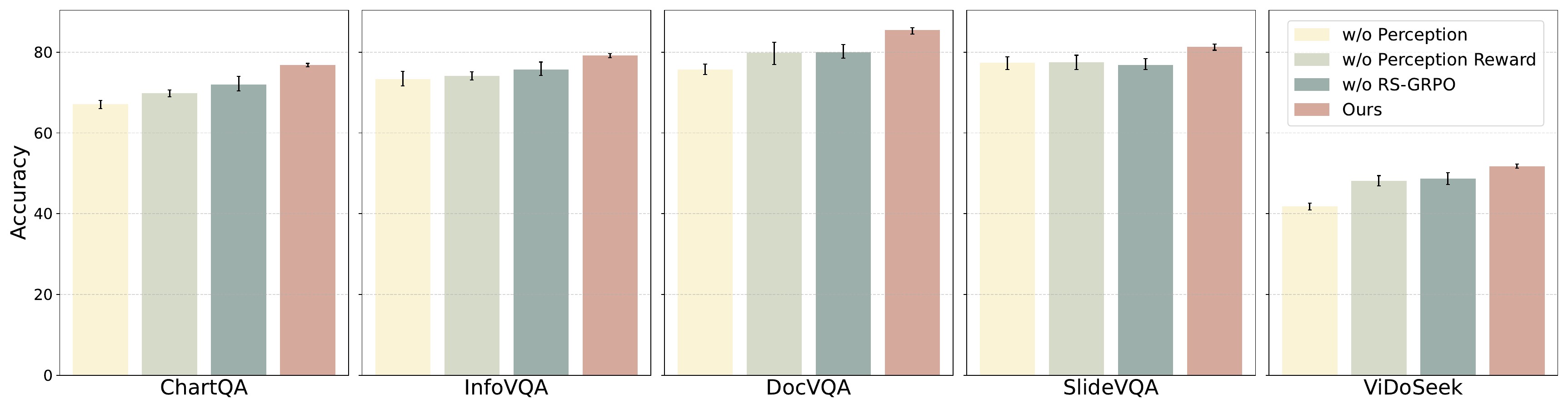}  
   \caption{Ablation study(\%): ``w/o Perception" trains the model with a standard think-then-answer approach on the same data. ``w/o Perception Reward" uses only answer correctness as the reward, omitting the additional Perception Reward. ``w/o RS-GRPO" sums all rewards and applies them to every token, corresponding to the standard GRPO algorithm. Results are averaged over 5 runs with different random seeds, and error bars indicate 95\% bootstrap confidence intervals.}
  \label{fig:ablation}
\end{figure*}

Figure~\ref{fig:ablation} presents an extended visualization of the ablation study in
Table~\ref{tab:ablation}, where we report \emph{95\% bootstrap confidence intervals}
computed from five runs using different random seeds. The intervals are estimated via
$10{,}000$ bootstrap resamples for each method–dataset pair, providing a more reliable
characterization of uncertainty compared to reporting only the mean and standard deviation.

Across all five benchmarks, the full EVisRAG model consistently achieves the highest
accuracy, with its confidence intervals being well separated from those of the ablated
variants in nearly all cases. This non–overlapping behavior indicates that the performance
gains from Perception modeling, Perception Reward, and RS-GRPO are statistically
significant rather than fluctuations due to random initialization. Moreover, the bootstrap
intervals of our complete method are noticeably narrower, demonstrating more stable
optimization dynamics. In contrast, removing any of the proposed components not only
reduces accuracy but also increases variance, highlighting the necessity and robustness
of each part of our reward design and training paradigm.

\subsection{Visual Attention Cases of EVisRAG}
We present in Figure~\ref{fig:appd_attn_case} a qualitative comparison of attention alignment with question-relevant visual evidence. The query asks: When was the article ``Why Young Americans are Driving So Much Less Than Their Parents" published? A human reader would first attend to the headline to verify topical relevance, then shift gaze to the metadata directly beneath it, where the publication date ``APR 10, 2012" appears. As shown in the figure, EVisRAG places greater attention mass on these evidence regions than the baselines, and in its reasoning trace, explicitly observes and records the date "APR 10, 2012," yielding the correct answer. This case illustrates that EVisRAG enhances perception during reasoning by aligning attention with task-critical visual evidence.

\subsection{Experiments on Natural Images}
\begin{table*}[t]
\centering
\setlength{\tabcolsep}{18pt}
\caption{Performance on natural-image QA.}
\begin{tabular}{l ccc ccc}
\toprule
& \multicolumn{3}{c}{\textbf{In Distribution} (all images as input)} 
& \multicolumn{2}{c}{\textbf{Out of Distribution} (top-3 recall)} \\
\cmidrule(lr){2-4} \cmidrule(lr){5-6}
& 2 images & 3 images & 5 images & 10 images & 50 images \\
\midrule
Qwen7b  & 64.67 & 64.44 & 62.00 & 57.8 & 56.2 \\
EVisRAG(Ours) & \textbf{86.22} \textcolor{green}{\scriptsize +21.55} & \textbf{85.33} \textcolor{green}{\scriptsize +20.89} & \textbf{83.11} \textcolor{green}{\scriptsize +21.11} & \textbf{71.6} \textcolor{green}{\scriptsize +13.8} & \textbf{64.5} \textcolor{green}{\scriptsize +8.3} \\
\bottomrule
\end{tabular}
\label{tab:nature}
\end{table*}

To further assess the generalizability of our method beyond document images, we additionally evaluate it on natural-image retrieval and reasoning under large-scale settings. We adopt the Visual Haystacks dataset~\citep{wu2024visual} as both training and evaluation data. From the 2-image, 3-image, and 5-image configurations, we first select the same 100 questions for each setting, resulting in 300 training examples in total, and use the remaining 900 questions in each configuration as test data. In addition, we evaluate on the 10-image and 50-image configurations, using all 1{,}000 questions in each as test examples. For the 10- and 50-image settings, we employ \texttt{clip-vit-large-patch14-336}\citep{radford2021learning} to retrieve the top-3 most relevant images, which are then fed into the model.

We compare our trained model against the original \textit{Qwen7B} model as the baseline. As shown in Table~\ref{tab:nature}, our approach achieves more than 20\% absolute accuracy improvement in the in-distribution settings (2, 3, and 5 images). In the out-of-distribution settings (10 and 50 images), even when relying on a relatively small CLIP model with imperfect retrieval quality, our method still yields on average more than 10\% absolute improvement. These results demonstrate that our approach remains highly effective on natural-image tasks and is not limited to document-centric scenarios.

\subsection{Efficiency under Larger Retrieval Contexts}
To further examine scalability with respect to retrieval size, we extend the number of retrieved images from Top-5 to Top-20 and report both effectiveness and efficiency.
As shown in Figure~\ref{fig:topk_eff}, EVisRAG consistently achieves higher accuracy than Qwen2.5-VL and R1-Router across all retrieval sizes, while using substantially fewer tokens than R1-Router. Although latency naturally increases with more retrieved images, EVisRAG maintains a favorable accuracy-efficiency trade-off in long-context multi-image settings.


\begin{figure}[t!]
  \centering
  \includegraphics[width=1.0\linewidth]{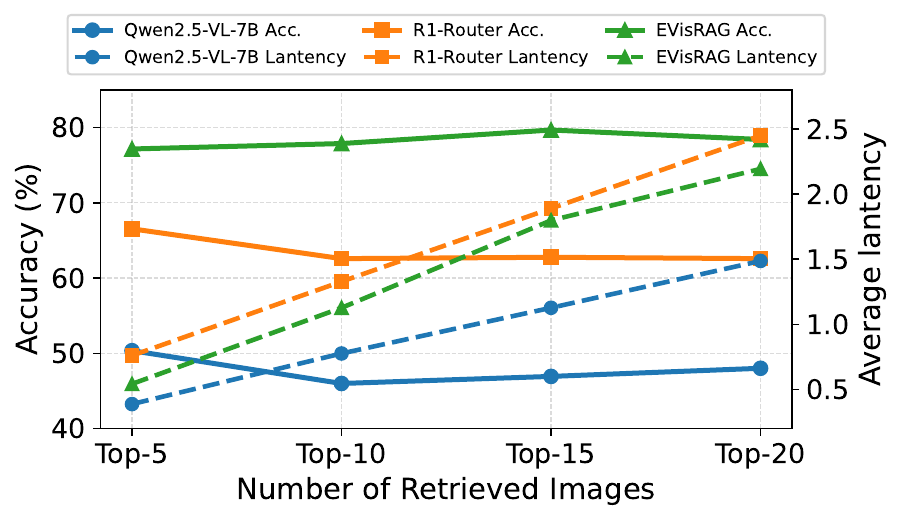}  
  \caption{Accuracy and output tokens under larger retrieval contexts.}
  \label{fig:retrieval}
\end{figure}

\subsection{Robustness to Different Retrievers}
\begin{table*}[t]
    \centering
    \setlength{\tabcolsep}{14pt}
    \caption{Performance of EVisRAG and Qwen7b with different retrievers on ViDoSeek.}
    \begin{tabular}{lccccc}
        \toprule
        & Sufficient Ratio & qwen7b-Acc & qwen7b-F1 & evisrag-Acc & evisrag-F1 \\
        \midrule
        VisRAG-ret~\citep{yu2024visrag}        & 84.24 & 42.56 & 42.48 & 52.10 \textcolor{green}{\scriptsize +9.5} & 65.78 \textcolor{green}{\scriptsize +23.3} \\
        Colpali-v1.3~\citep{faysse2024colpaliefficientdocumentretrieval}        & 84.33 & 42.23 & 40.95 & 50.79 \textcolor{green}{\scriptsize +8.6} & 63.82 \textcolor{green}{\scriptsize +22.9} \\
        jina-embeddings-v4~\citep{günther2025jinaembeddingsv4universalembeddingsmultimodal}   & 85.11 & 40.72 & 53.02 & 49.37 \textcolor{green}{\scriptsize +8.7} & 64.01 \textcolor{green}{\scriptsize +11.0} \\
        \bottomrule
    \end{tabular}
    \label{tab:retrive}
\end{table*}

To examine whether our approach depends on a specific retrieval module, we further evaluate the trained model under multiple independent retrievers. Although our method is trained with VisRAG-Ret as the retrieval component, at inference time, we replace the retriever with two alternative models of different architectures and scales: Colpali-v1.3 and Jina-embeddings-v4. For each retriever, we obtain the top-3 relevant images and feed them into our QA model without any retraining or adaptation. The results in Table~\ref{tab:retrive} demonstrate that our method yields consistent and substantial improvements across all retrievers, which confirms that our approach is retriever-agnostic.

\subsection{Case Studies of EVisRAG}

\begin{figure*}[t!]
  \centering
  \includegraphics[width=0.75\textwidth]{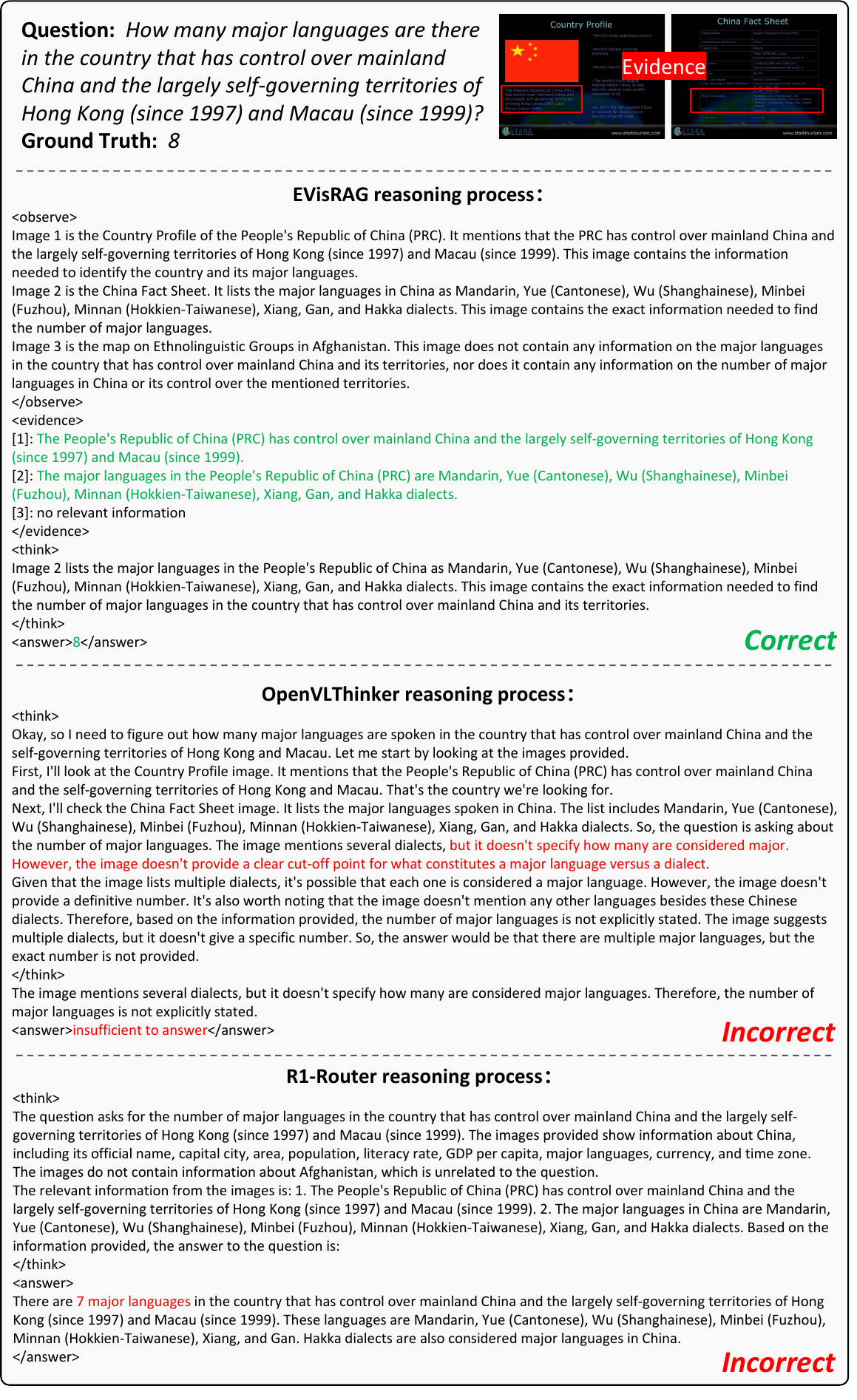}  
  \caption{A Case Study on SlideVQA}
  \label{fig:case2}
\end{figure*}

In this section, we analyze a multi-hop case in Figure~\ref{fig:case2} from the SlideVQA dataset. The question asks for the number of major languages in the country that governs mainland China and the largely self-governing territories of Hong Kong (since 1997) and Macau (since 1999). Answering requires integrating evidence from two slides: one identifies the country as China. The other enumerates China’s major languages, including Mandarin, Yue (Cantonese), Wu (Shanghainese), Minbei (Fuzhou), Minnan (Hokkien–Taiwanese), Xiang, Gan, and Hakka, a total of eight. EVisRAG correctly records the provenance of each piece of evidence and produces the correct answer, demonstrating both reliable visual perception and cross-page reasoning. In contrast, OpenVLThinker and R1-Router fail: OpenVLThinker infers the correct subgoal but, having missed the second slide’s list, predicts that no answer exists. R1-Router locates both slides but misperceives the list and counts seven instead of eight.


\subsection{License}
We strictly comply with the original licenses and release terms of all datasets used in this work and do not redistribute any third-party raw images or proprietary data. All datasets are used solely for research and evaluation purposes. For datasets without explicitly stated open licenses, we follow their original release conditions, restrict usage to non-commercial academic research, and do not redistribute any raw data or images. All released artifacts (including code, model checkpoints, and processed metadata) exclude any proprietary or restricted content.

\begin{figure*}[htbp]
  \centering
  \includegraphics[width=1.0\textwidth]{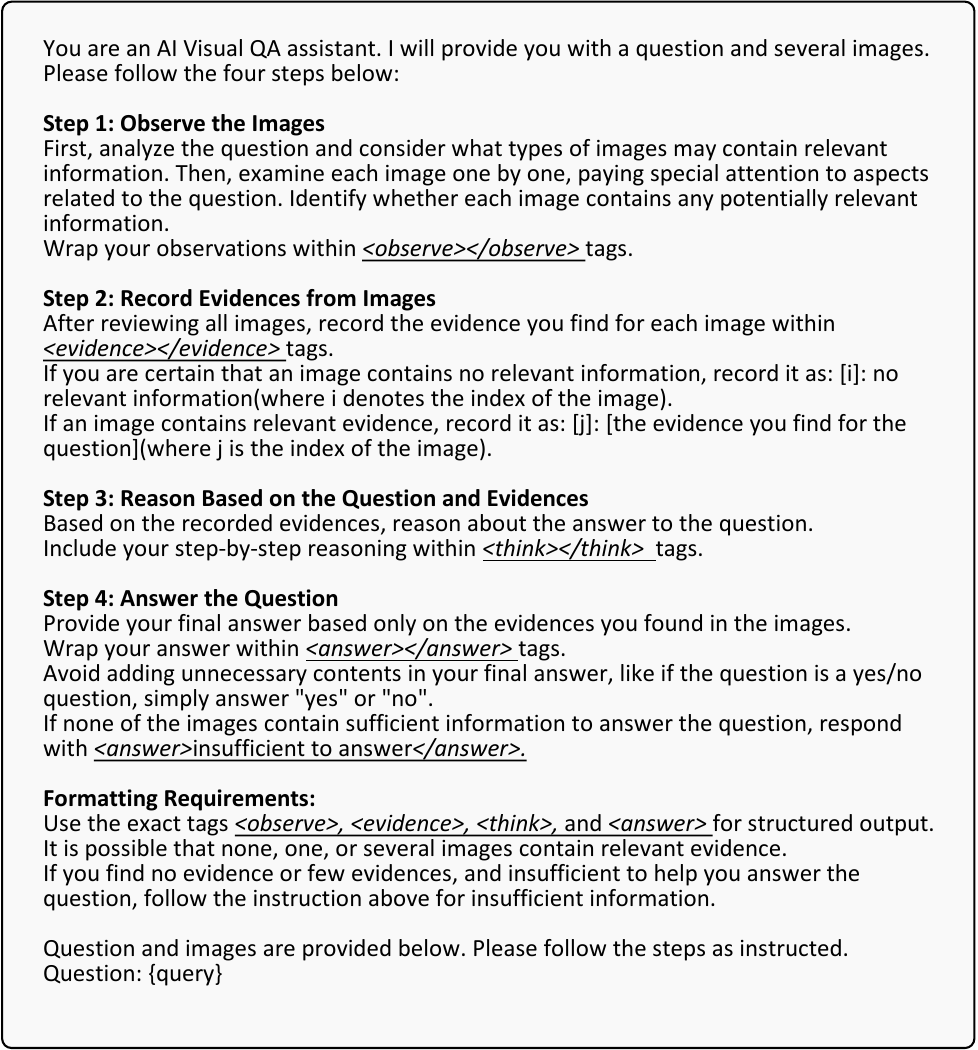}  
  \caption{The Prompt Template for EVisRAG(SFT\&GRPO)}
  \label{fig:.evidencetrain_prompt}
\end{figure*}


\end{document}